\documentclass[11pt]{article}

\usepackage[final]{acl}

\usepackage{times}
\usepackage{latexsym}

\usepackage[T1]{fontenc}

\usepackage[utf8]{inputenc}

\usepackage{microtype}

\usepackage{inconsolata}

\usepackage{algorithm}
\usepackage{algpseudocode}
\usepackage{amsmath}
\usepackage{amssymb}

\usepackage{xcolor}

\hypersetup{
    colorlinks=true,
    urlcolor=blue
}
\usepackage{enumitem}

\usepackage[table]{xcolor}
\usepackage{booktabs}
\usepackage{array}
\usepackage{soul}
\usepackage{graphicx}
\newcolumntype{L}[1]{>{\raggedright\arraybackslash}p{#1}}

\renewcommand{\arraystretch}{1.08}
\title{Look Before You Leap: Factual Decoding with Internal\\ Attribution Signals}

\author{
Hayeong Ryu$^{\mathbf{1}}$,
JungMin Yun$^{\mathbf{1}}$,
Byeonggeuk Lim$^{\mathbf{2}}$,
Sunhee Jo$^{\mathbf{2}}$,
YoungBin Kim$^{\mathbf{1,2}}$
\\
$^{1}$Department of Artificial Intelligence, Chung-Ang University
\\
$^{2}$Graduate School of Advanced Imaging Sciences, Multimedia and Film, Chung-Ang University
\\
{\ttfamily \{bluebarry37, cocoro357, banggeuk, jo3438, ybkim85\}@cau.ac.kr}
}

\begin{document}
\maketitle
\begin{abstract}
    Hallucination remains a critical challenge in large language models (LLMs), 
    where early factual errors compound through autoregressive generation in a 
    snowballing effect that neither post-hoc correction nor weight-level 
    intervention can effectively preempt. We propose \textbf{\textsc{DescaPE}} 
    (\textbf{DE}coding \textbf{S}ignal \textbf{C}ontrol \textbf{A}gainst 
    \textbf{P}ath \textbf{E}rror-snowballing), a decoding framework that 
    leverages internal model signals to suppress hallucination-prone trajectories 
    at inference time. Through sliding-window MLP ablation, we identify a 
    factual-salient layer span within LLMs whose derived signal is 
    selectively elevated for factual tokens and exhibits anomalous spikes at 
    hallucination-prone steps. We train a lightweight probe to approximate this 
    signal from a single forward pass and integrate it into candidate scoring to 
    penalize high-risk continuations while rewarding factually grounded ones. 
    Experiments across five factuality benchmarks on three LLMs demonstrate that 
    \textsc{DescaPE} achieves factuality improvements over decoding-time baselines in multiple settings, while incurring only 1.10$\times$ latency overhead in our efficiency evaluation.
    Our code is available at \url{https://github.com/hayeonggg/DESCAPE}.
\end{abstract}

\section{Introduction}
\label{introduction}
Despite remarkable advances in large language models (LLMs), hallucination remains a critical and unresolved challenge~\cite{03,04,10,01,02}. Retrieval-Augmented Generation (RAG)~\cite{05}, Supervised Fine-Tuning (SFT)~\cite{06}, and Reinforcement Learning from Human Feedback (RLHF)~\cite{07} have been widely adopted to mitigate this issue, yet each introduces fundamental limitations: RAG is structurally dependent on retriever quality and incurs search latency, while SFT and RLHF demand substantial computational cost and risk catastrophic forgetting. More critically, the autoregressive generation process allows early factual errors to compound and propagate, a phenomenon known as hallucination snowballing~\cite{08}, which neither post-hoc corrections nor weight-level interventions can effectively preempt~\cite{09}. This underscores the need for decoding-time intervention: a lightweight mechanism that suppresses factual errors during generation, without modifying model parameters or relying on external modules.

\begin{figure}[t!]
    \centering
    \includegraphics[width=\columnwidth,clip,trim=5 5 5 5]{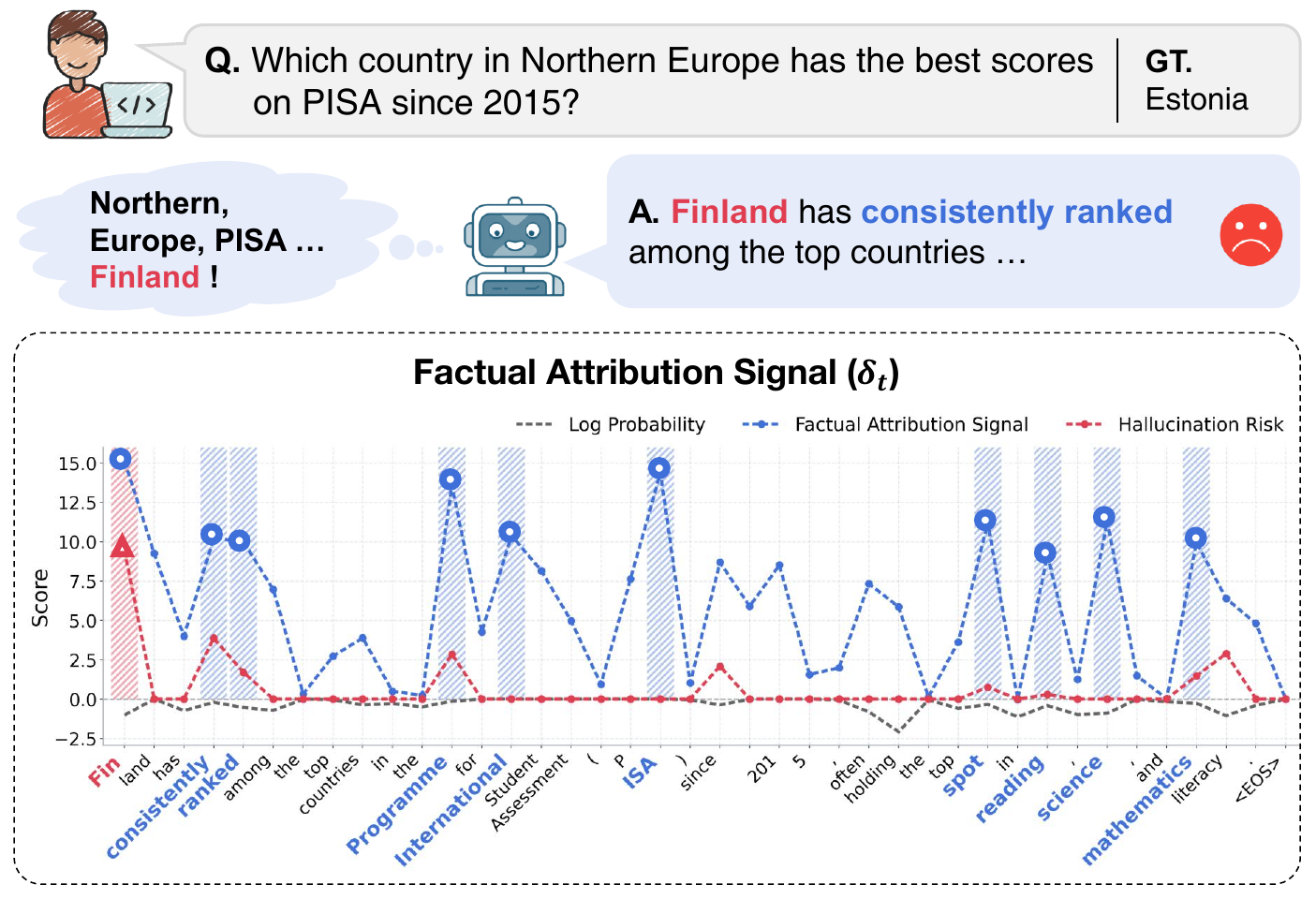}
    \caption{Illustration of our preliminary analysis motivating the proposed method.
    The Factual Attribution Signal $\delta_t$, extracted from the factual-salient layer span,
    selectively responds to factual content tokens (\textbf{\textcolor{blue}{blue}}) throughout the generated sequence.
    Simultaneously, when the model generates a hallucinated token (\textit{Finland}, \textbf{\textcolor{red}{red}}) with misplaced confidence,
    $\delta_t$ exhibits an anomalous spike---triggering a risk signal that can prevent hallucination snowballing before it propagates.}
    \vspace{-10pt}
    \label{fig:overview}
\end{figure}


Decoding-time factuality research has evolved along three paradigms. Intrinsic Steering methods ~\cite{11,13,12} steer internal representations toward truthful outputs, but offer limited recourse once a hallucinated trajectory is initiated. Extrinsic Verification methods ~\cite{14,15,16} rescore candidates via auxiliary signals at the cost of inference overhead and distribution shift. Self-Verification methods ~\cite{17,18} leverage self-feedback without external verifiers, yet remain fundamentally bounded by the generator's own distributional space. This raises a fundamental question: whether an independent factual signal --- internal to the model yet decoupled from its final output distribution --- can exist.

\begin{figure*}[t!]
    \centering
    \includegraphics[width=1.0\textwidth]{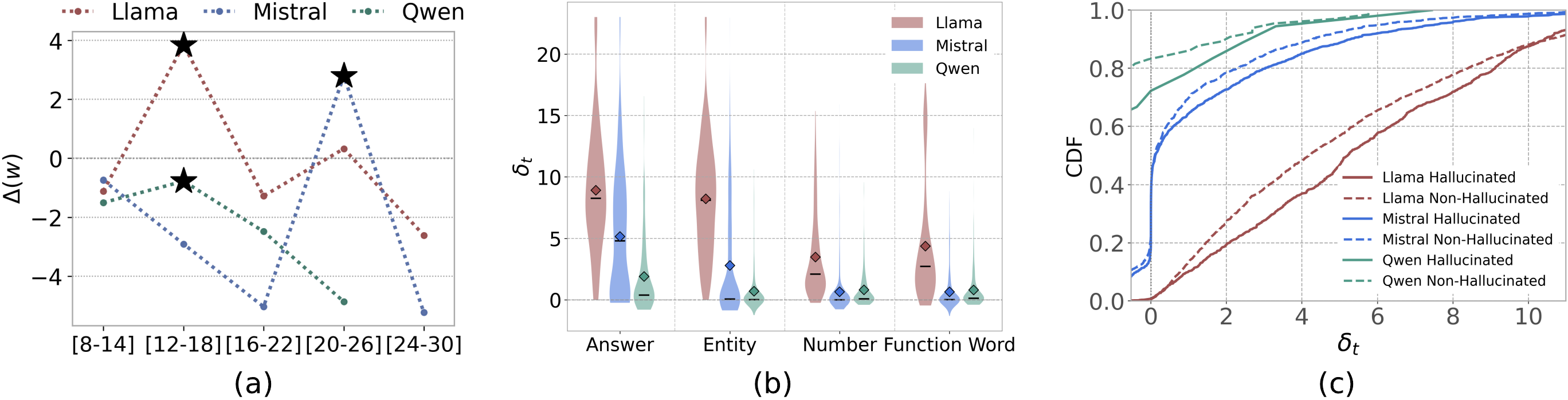}
    \caption{Analysis of factual-salient layer spans across three models. \textbf{(a)} Factual Attribution Score $\Delta(w)$ for each window, calculated using sliding-window MLP ablation. The $\bigstar$ symbol indicates the factual-salient layer span identified for each model. \textbf{(b)} Distribution of $\delta_t$ derived from the identified spans, categorized by token type. \textbf{(c)} Empirical CDF of $\delta_t$ for hallucinated (solid line) and non-hallucinated (dashed line) tokens.}
    \label{fig:observation}
    \vspace{-4pt}
\end{figure*}

Recent work reports that LLMs exhibit a late sharpening phenomenon~\cite{19,20}, where upper layers sharply narrow probability distributions, suggesting that erroneous trajectories may be internally set before a hallucinated token is emitted~\cite{11,12}. Yet, this observation paradoxically implies that intermediate layers, prior to this sharpening, harbor independent factual signals, making preemptive decoding-time intervention structurally plausible. 

Motivated by recent causal analyses of factual recall~\cite{21}, we analyze layer-wise MLP activations during text generation and present two key findings. (i) We isolate a contiguous factual-salient layer span selectively engaged during factual generation: the derived factual attribution signal is consistently elevated for knowledge-intensive tokens while remaining negligible for functional tokens. (ii) Hallucinated tokens trigger abnormally frequent spikes in this signal relative to factual ones, establishing it as a reliable signal for identifying hallucination-prone continuations. As shown in Figure~\ref{fig:overview}, this signal selectively tracks factual content throughout generation and exhibits anomalous spikes at hallucination-prone steps, providing a real-time basis for preemptive intervention.

In this work, we propose \textbf{DE}coding \textbf{S}ignal \textbf{C}ontrol \textbf{A}gainst \textbf{P}ath \textbf{E}rror-snowballing (\textsc{DescaPE}), a decoding framework that suppresses hallucination-prone trajectories at inference time. We train a lightweight probe to approximate the factual attribution signal from a single forward pass and integrate it into candidate scoring to penalize high-risk continuations while favoring factually grounded ones. Experiments across five factuality benchmarks show that \textsc{DescaPE} improves factuality in several settings with minimal inference overhead.
Our contributions are threefold:
\begin{enumerate}[noitemsep, topsep=2pt]
    \item We identify a factual-salient layer span within LLMs whose derived attribution signal is selectively elevated for factual tokens, and further demonstrate that it exhibits anomalous spikes at hallucination-prone generation steps.
    \item We propose \textsc{DescaPE}, a preventive trajectory control framework that leverages this signal as a unified internal basis for decoding-time intervention, approximating it via a lightweight probe and integrating it into candidate scoring.
    \item We validate \textsc{DescaPE} across five factuality benchmarks, demonstrating improvements in several settings with minimal latency overhead.
\end{enumerate}









\section{Related Work}
\label{related_work}

\subsection{Intrinsic Signal-Based Intervention}
Prior work has explored improving factual generation by directly manipulating internal representations or layer-wise signals of LLMs. Some intervene at the representation level by steering activations toward truthful directions (ITI~\cite{34}; TrFr~\cite{35}; TruthX~\cite{36}). Others exploit layer-wise distributional differences at decoding time, contrasting or reweighting logits across layers to promote factual predictions (DoLa~\cite{11}; SLED~\cite{13}; END~\cite{33}; ActLCD~\cite{12}). 
However, these layer-wise adjustments are uniformly applied across generation steps, without explicit awareness of whether the current step is hallucination-prone. \textsc{DescaPE} derives a candidate-level attribution signal from a causally identified factual-salient layer span, enabling selective intervention at hallucination-prone steps.

\subsection{Extrinsic Verification Decoding}
Another approach rescores or filters candidate continuations using signals from auxiliary models external to the generator. Contrastive Decoding~\cite{14} amplifies the logit gap between a strong and weak model, with subsequent work refining the contrastive signal via hallucination-induced fine-tuning~\cite{38} and adversarial hard negatives~\cite{39}. Monitoring Decoding~\cite{15} further advances toward token-level intervention by deploying a large external monitor to resample high-risk tokens in real time. Despite their precision, these methods introduce distribution mismatch and non-trivial inference overhead — limitations our approach overcomes by deriving an equivalent signal entirely from the generator's own factual-salient layer span.

\subsection{Self-Verification Decoding}
Self-verification methods improve factual generation by having the model evaluate and revise its own outputs without external models. Sampling-based approaches exploit cross-sample agreement as a proxy for factual reliability~\cite{41,42,43}, while recent work extends this to decoding time: Integrative Decoding~\cite{18} incorporates self-consistency into the decoding objective, and DSVD~\cite{17} triggers dynamic rollback upon erroneous token detection. However, since verification and correction operate within the same representation space, the model is structurally ill-equipped to escape incorrect convictions~\cite{47}, and post-detection intervention cannot preempt erroneous trajectories. Our framework addresses both by deriving signals from a factual-salient layer span distinct from the final output distribution, enabling preventive trajectory control.

\section{ Identifying Factual-Salient Layer Spans}\label{3}
\label{observation}
To proactively prevent entering hallucination-prone trajectories at the decoding stage, we require an internal signal that can be utilized in real time during generation without modifying the base model parameters or relying on an external verifier LLM. We therefore identify a contiguous span of MLP layers that is selectively engaged in factual generation — which we term the factual-salient layer span — and demonstrate that the signal derived from this span serves as an early indicator of hallucination.

\subsection{Background and Motivation} \label{3.1}
Prior work has established through causal tracing that the recall of factual knowledge in LLMs is causally localized in mid-layer MLP modules~\cite{21}, with MLP sublayers playing a central role in enriching subject token representations and retrieving factual attributes~\cite{22}. Building on this evidence, we focus on contiguous MLP spans that selectively contribute to factual token prediction and use their causal contribution as a candidate-level attribution signal for decoding-time intervention. Component-wise ablation further supports this choice, with MLPs exhibiting
stronger factual attribution and factual--function separation than attention modules (Appendix~\ref{appendix_component_ablation}).

\begin{figure*}[t!]
    \centering
    \includegraphics[width=1.0\textwidth]{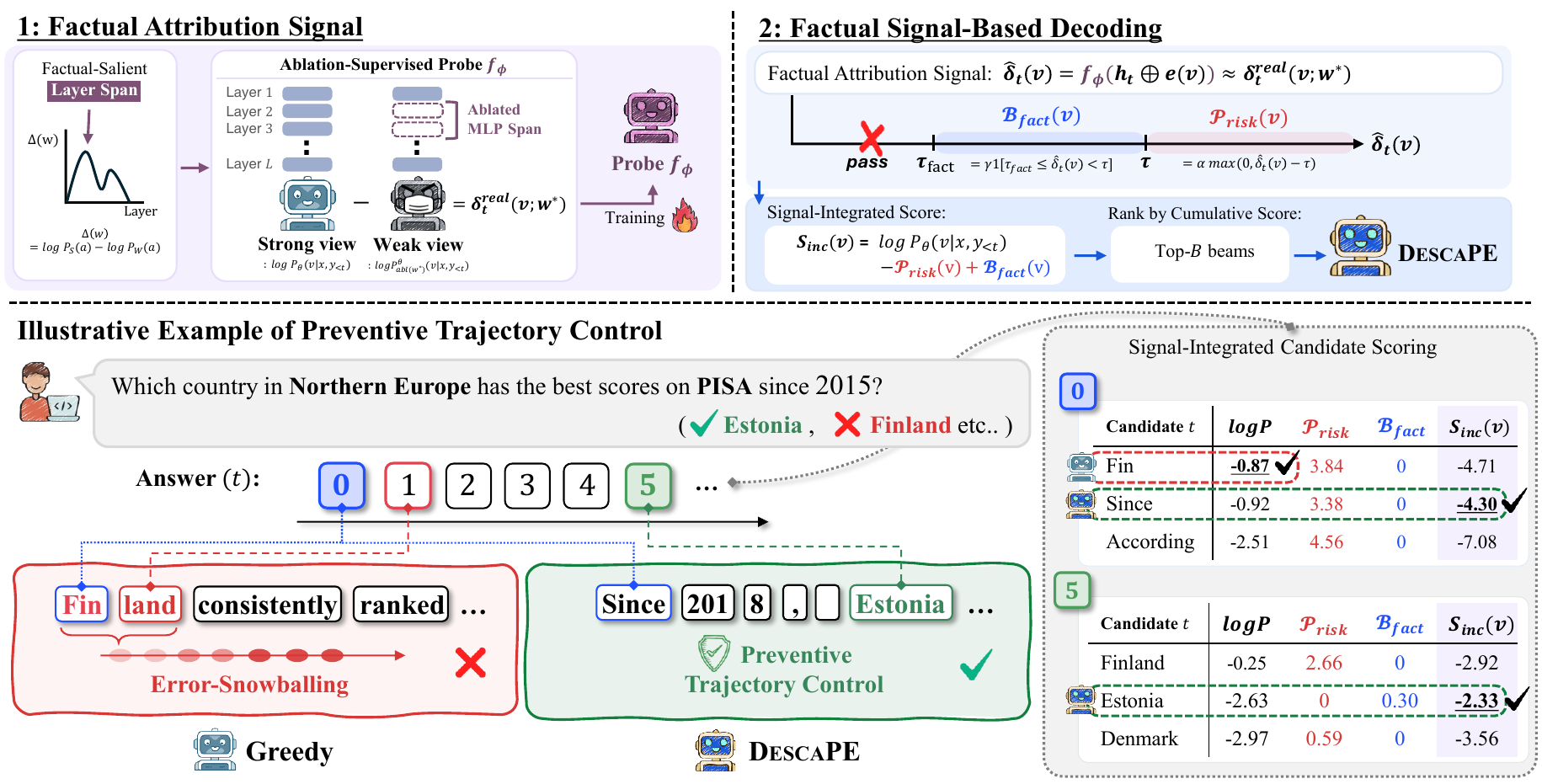}
    \caption{
    Overview of \textbf{\textsc{DescaPE}}.
    \textbf{(Top-left)} A lightweight probe $f_\phi$ approximates the
    candidate-level factual attribution signal $\hat{\delta}_t(v)$ using
    supervision derived from the factual-salient layer span identified via
    sliding-window ablation.
    \textbf{(Top-right)} The estimated signal is incorporated into candidate
    scoring through a risk penalty $\mathcal{P}_{\mathrm{risk}}$ and factual
    bonus $\mathcal{B}_{\mathrm{fact}}$ to guide beam selection.
    \textbf{(Bottom)} An illustrative decoding example shows how this
    intervention prevents error snowballing. At $t=0$, the base model assigns
    higher probability to the hallucination-prone candidate \textit{Finland},
    whereas \textsc{DescaPE} penalizes it according to its high attribution
    signal and promotes the factually grounded candidate \textit{Estonia}.
    This redirects generation away from the erroneous greedy trajectory, with
    signal-integrated scoring at subsequent steps maintaining the corrected
    trajectory toward a factually grounded response.
    }
    \vspace{-8pt}
    \label{fig:main}
\end{figure*}

\subsection{Sliding-Window MLP Ablation} \label{3.2}
To quantify the contribution of each layer span to factual token generation, we propose a sliding-window MLP ablation approach. Specifically, we define a contiguous layer span $[l_{\text{start}}, l_{\text{end}}]$ as a single window and block the information flow by zeroing out the output activations of all MLP modules within this span. We quantify the causal contribution of each window $w$ using the Factual Attribution Score $\Delta(w)$ as follows:
\begin{equation}
    \Delta(w) = \log P_{S}(a) - \log P_{W}(a),
    \label{eq1}
\end{equation}
where $P_{S}(a)$ is the probability of the original, un-intervened model (Strong view) predicting the target answer token $a$, and $P_{W}(a)$ is the probability of the ablated model (Weak view) generating the same token $a$ with window $w$ zeroed out at the corresponding decoding step. A higher $\Delta(w)$ indicates a greater contribution of the span to factual token generation. For multi-token answers, we average the $\Delta(w)$ values across all answer tokens to compute the final score for the sample. 
We designate the span that yields the maximum $\Delta(w)$ as the \textbf{factual-salient layer span}, denoted by $w^* = \arg\max_{w} \Delta(w)$. Given this fixed span, at decoding step $t$, we define the real token-level factual attribution signal for a candidate token $v$ as
\begin{equation}
\begin{aligned}
    \delta_t^{\mathrm{real}}(v; w^*)
    &=
    \log p_{\theta}(v \mid x, y_{<t}) \\
    &\quad -
    \log p_{\theta_{\mathrm{abl}}(w^*)}(v \mid x, y_{<t}),
\end{aligned}
\label{eq:delta_real}
\end{equation}
where $p_{\theta}^{\mathrm{abl}(w^*)}(v \mid x, y_{<t})$ denotes the probability assigned to $v$ by the same model when the MLP outputs in $w^{*}$ are zero-ablated. The resulting $\delta_t^{\mathrm{real}}$ represents the attribution signal obtained through actual ablation, rather than a factuality label.
More detail is in Appendix~\ref{appendix_A_data}.

\subsection{Empirical Observations on Factual Signal} \label{3.3}
We apply the proposed sliding-window approach to three models: Llama-3.1-8B-Instruct~\cite{29}, Mistral-7B-Instruct-v0.3~\cite{30}, and Qwen2.5-7B-Instruct~\cite{31}. As shown in Figure~\ref{fig:observation} (a), all models exhibit a pronounced peak in $\Delta(w)$ at a specific layer span, confirming that a contiguous region of MLP layers is selectively engaged in factual generation. Specifically, the peak is observed at layers 12–18 for Llama, layers 20–26 for Mistral, and layers 12–18 for Qwen\footnote{For Qwen, the signal is more diffuse across spans, though we consistently designate the span with the highest $\Delta(w)$ as the factual-salient layer span; the validity of this choice is further corroborated in Section~\ref{5.2}.}. 
We then analyze the generated-token signal $\delta_t := \delta_t^{\mathrm{real}}(y_t; w^*)$, where $y_t$ denotes the token generated at decoding step $t$, and make two key observations.

\paragraph{Observation 1: The signal $\delta_t$ is selectively activated by factual tokens.}
Across all three models, $\delta_t$ is consistently elevated for factual tokens (e.g., answers, entities, numbers), while remaining near zero for function words (Figure~\ref{fig:observation} (b)). This indicates that the factual-salient layer span activates selectively during factual knowledge retrieval, with minimal response to function words.

\paragraph{Observation 2: Hallucinated tokens induce abnormal spikes in $\delta_t$.}
We further find that hallucinated tokens exhibit more frequent abnormal spikes in $\delta_t$ compared to non-hallucinated tokens. As shown in Figure~\ref{fig:observation} (c), the empirical Cumulative Distribution Function (CDF) of non-hallucinated tokens (dashed) lies consistently above that of hallucinated tokens (solid) across all models, even as $\delta_t$ remains broadly elevated at factual generation steps. We leverage this signal as a risk indicator in our decoding framework\footnote{
Although Figure~2 analyzes the signal on generated tokens, Eq.~\ref{eq:delta_real} defines $\delta_t^{\mathrm{real}}(v; w^*)$ for candidate tokens prior to token selection.
We further examine whether the resulting intervention preserves correct multi-hop reasoning on HotpotQA; see Appendix~\ref{appendix_reasoning_preservation}.
}; further details are provided in Appendix~\ref{appendix_B_analysis}.

\section{Methodology}
\label{method}
In this section, we introduce \textbf{DE}coding \textbf{S}ignal \textbf{C}ontrol 
\textbf{A}gainst \textbf{P}ath \textbf{E}rror-snowballing (\textsc{DescaPE}), 
a decoding framework that leverages internal signals from the factual-salient layer span 
(Section~\ref{3}) to suppress hallucination-prone trajectories at inference time. 
It consists of (1) a lightweight probe that approximates the factual attribution signal 
from a single forward pass, and (2) a signal-integrated beam scoring strategy that 
penalizes high-risk continuations while rewarding factually grounded ones. 
An overview is illustrated in Figure~\ref{fig:main}.

\subsection{Factual Attribution Signal Approximation} \label{4.1}
While our factual attribution signal $\delta_t^{\mathrm{real}}(v; w^*)$ can be obtained by applying Eq.~\ref{eq:delta_real} at each decoding step, performing dual forward passes at every step incurs substantial computational overhead. We therefore propose to train a lightweight probing head that approximates this signal from the hidden representations of a single forward pass, using $\delta_t^{\mathrm{real}}(v; w^*)$  as the supervision target. 

\paragraph{Training Data Construction.}
To collect supervision targets, we run autoregressive generation on up to 3,000 samples from Databricks Dolly-15k\footnote{\url{https://huggingface.co/datasets/databricks/databricks-dolly-15k}} and record $\delta_t^{\mathrm{real}}(v; w^*)$ for each candidate token at every decoding step. Further details on the data collection procedure are provided in Appendix~\ref{appendix_C_data}.

\paragraph{Probe Architecture and Training.}
To capture the sparse yet decisive spike patterns identified in Section~\ref{3.3} (Observation 2), we extend the Huber loss with spike-sensitive weights:
\begin{equation}
    \mathcal{L}
    =
    \frac{1}{N}
    \sum_{i=1}^{N}
    \omega_i
    \cdot
    \mathrm{Huber}(\hat{\delta}_i, \delta_i^{\mathrm{real}}, \epsilon_h),
    \label{eq2}
\end{equation}
where $\hat{\delta}_i$ is the probe's prediction, $\epsilon_h$ is the Huber threshold, and $\omega_i = \mathrm{softmax}(\beta \cdot \max(0,\delta_i^{\mathrm{real}})) \cdot N$ dynamically up-weights samples with large positive supervision targets, encouraging the probe to focus on high-spike regions. More training details are provided in Appendix~\ref{appendix_D_probe}.


\definecolor{rowblue}{HTML}{E8F4FA}

\newcommand{\best}[1]{\textbf{#1}}
\newcommand{\second}[1]{\underline{#1}}

\begin{table*}[t]
\centering
\setlength{\tabcolsep}{3.8pt}
\renewcommand{\arraystretch}{1.00}
\footnotesize

\resizebox{0.9\textwidth}{!}{%
\begin{tabular}{l ccc !{\vrule width 0.4pt} cc !{\vrule width 0.4pt} c !{\vrule width 0.4pt} cc !{\vrule width 0.4pt} cc}
\toprule
\raisebox{-2.2ex}[0pt][0pt]{\textbf{Method}} &
\multicolumn{3}{c}{\textbf{TruthfulQA}} &
\multicolumn{2}{c}{\textbf{FreshQA}} &
\multicolumn{1}{c}{\textbf{\textsc{FActScore}}} &
\multicolumn{2}{c}{\textbf{NQ}} &
\multicolumn{2}{c}{\textbf{TriviaQA}} \\
\cmidrule(lr){2-11}
& Truth(\%) & Info(\%) & T*I(\%) & Strict & Relaxed & Score & EM & F1 & EM & F1 \\
\midrule

Llama-3.1      & \second{62.9} & 69.9 & 44.4 & \second{30.0} & 33.0 & 47.10 & \best{10.1} & \best{25.2} & 35.7 & 54.6 \\
+ Beam Search  & \best{63.6} & 71.3 & 44.9 & 28.8 & \best{33.8} & 44.90 & 7.7 & \second{24.4} & 53.6 & \second{58.5} \\
+ Self-Refine  & 61.8 & \second{75.8} & \second{46.6} & 27.7 & 31.7 & \second{50.72} & 6.7 & 20.9 & \second{54.6} & 55.1 \\
+ DoLa         & 57.2 & 65.0 & 39.3 & 16.8 & 25.2 & 47.26 & \second{8.9} & 21.6 & 47.3 & 49.4 \\
+ SLED         & 57.1 & 72.9 & 42.5 & 15.7 & 27.5 & 49.30 & 4.0 & 16.4 & 42.7 & 56.6 \\
+ END          & 60.5 & 74.2 & 44.9 & 13.8 & 13.0 & 37.23 & 1.7 & 8.5 & 24.6 & 35.4 \\
+ ActLCD       & 61.5 & 72.3 & 45.1 & \best{31.0} & 32.2 & 43.20 & 7.7 & 22.3 & \best{59.8} & \best{69.3} \\
\rowcolor{rowblue}
+ \textbf{\textsc{DescaPE}} (ours) & 62.0 & \best{78.2} & \best{48.5} & 29.2 & \second{33.2} & \best{67.20} & 1.8 & 23.4 & 39.8 & 55.6 \\

\midrule

Mistral-v0.3   & 66.3 & 62.3 & 41.6 & \second{26.3} & 33.3 & 47.0 & 0.3 & \second{14.9} & \second{35.2} & 54.4 \\
+ Beam Search  & \second{66.9} & 63.2 & 42.6 & 25.7 & 32.8 & \best{48.2} & \second{0.4} & 13.7 & \best{41.2} & \second{58.4} \\
+ Self-Refine  & 66.1 & 66.6 & \second{43.3} & \best{26.5} & 32.5 & 47.8 & 0.1 & 12.7 & 26.1 & 46.8 \\
+ DoLa         & 65.9 & 65.5 & 43.0 & 26.0 & 33.2 & 45.6 & 0.2 & 14.7 & 30.7 & 51.8 \\
+ SLED         & 65.7 & \second{67.7} & \second{43.3} & 20.5 & 31.7 & 45.0 & 0.1 & 13.0 & 34.6 & 53.1 \\
+ END          & 66.4 & \best{67.9} & 43.1 & 12.0 & 12.5 & 21.6 & 0.3 & 4.5 & 13.5 & 25.7 \\
+ ActLCD       & 65.4 & 65.7 & 43.1 & 25.5 & \second{34.0} & 46.7 & 0.1 & 14.0 & 29.0 & 49.6 \\
\rowcolor{rowblue}
+ \textbf{\textsc{DescaPE}} (ours) & \best{67.6} & 65.3 & \best{43.4} & \best{26.5} & \best{34.8} & \second{47.9} & \best{0.5} & \best{15.7} & \best{41.2} & \best{58.8} \\

\midrule

Qwen2.5        & 65.0 & 60.9 & 39.7 & 24.3 & 29.8 & 40.7 & \second{11.1} & \second{22.9} & 48.1 & 56.0 \\
+ Beam Search  & 65.5 & 62.7 & 41.5 & \second{25.5} & \best{33.7} & \second{41.4} & 8.0 & 20.4 & \second{48.4} & \second{56.6} \\
+ Self-Refine  & 65.1 & 62.8 & 40.9 & 25.0 & 31.2 & \best{44.0} & 7.7 & 19.8 & 35.8 & 47.3 \\
+ DoLa         & 58.4 & 55.0 & 34.3 & 21.8 & 27.3 & 28.6 & 7.6 & 18.2 & 38.2 & 45.8 \\
+ SLED         & 64.2 & \best{63.7} & 41.7 & 22.0 & 30.5 & 39.1 & 9.3 & 20.9 & 44.2 & 52.6 \\
+ END          & \second{65.8} & \second{63.1} & \second{42.1} & 23.2 & 28.5 & 38.3 & 9.9 & 21.9 & 45.8 & 54.1 \\
+ ActLCD       & 65.2 & 60.9 & 40.0 & 24.8 & 30.5 & 40.9 & \best{11.4} & \best{23.1} & 47.9 & 56.0 \\
\rowcolor{rowblue}
+ \textbf{\textsc{DescaPE}} (ours) & \best{66.4} & 62.8 & \best{42.2} & \best{26.0} & \second{31.3} & 40.8 & 8.0 & 21.3 & \best{49.6} & \best{57.3} \\

\bottomrule
\end{tabular}%
}
\caption{Results across TruthfulQA, FreshQA, \textsc{FActScore}, NQ, and TriviaQA. Best results in each model block are shown in \textbf{bold}, and second-best results are \underline{underlined}.}
\vspace{-8pt} 
\label{tab:main_result}
\end{table*}

\subsection{Factual Signal-Based Decoding} \label{4.2}
By incorporating $\hat{\delta}_t(v)$ predicted by $f_\phi$ into the beam scoring at each decoding step, we directly leverage the factual attribution signal at inference time. Specifically, tokens exhibiting abnormally high attribution scores are penalized, while those falling within the stable factual zone receive a bonus, proactively blocking entry into hallucination-prone trajectories and actively steering generation toward factually grounded continuations.

\paragraph{Factual Attribution Signal.}
At each decoding step $t$, the probe approximates the real token-level factual attribution signal for each candidate token $v$ as:
\begin{equation}
    \hat{\delta}_t(v) = f_\phi(h_t \oplus e(v))
    \approx
    \delta_t^{\mathrm{real}}(v; w^*),
\label{eq3}
\end{equation}
where $h_t$ is the hidden state extracted from the last layer of the factual-salient layer span, which is concatenated with the embedding $e(v)$ of candidate token $v$ and passed through the probing head $f_\phi$ to produce the score.

\paragraph{Signal-Integrated Scoring.}
To suppress hallucination-prone continuations and promote factually grounded ones, we compute a risk penalty $\mathcal{P}_{\text{risk}}(v)$ and a factual bonus $\mathcal{B}_{\text{fact}}(v)$ for each candidate token:
\begin{equation}
    \mathcal{P}_{\text{risk}}(v) = \alpha \cdot \max(0, \hat{\delta}_t(v) - \tau),
\label{eq4}
\end{equation}
\begin{equation}
    \mathcal{B}_{\text{fact}}(v) = \gamma \cdot 1[\tau_{\text{fact}} \le \hat{\delta}_t(v) < \tau],
\label{eq5}
\end{equation}
where $\alpha$ and $\gamma$ are scalar weights controlling the strength of the penalty and bonus, respectively.
The thresholds $\tau$ and $\tau_{\mathrm{fact}}$ partition a signal value $s$ into three zones: safe ($s < \tau_{\mathrm{fact}}$), factual ($\tau_{\mathrm{fact}} \leq s < \tau$), and risk ($s \geq \tau$). $\mathcal{P}_{\text{risk}}(v)$ suppresses entry into high-risk continuations by penalizing tokens whose $\hat{\delta}_t(v)$ exceeds threshold $\tau$, corresponding to the anomalous attribution spikes observed at hallucination-prone steps. $\mathcal{B}_{\text{fact}}(v)$ rewards tokens falling within the stable factual attribution zone $[\tau_{\text{fact}}, \tau)$, actively encouraging the model to select factually supported tokens. Tokens below $\tau_{\text{fact}}$ are excluded from the bonus, as they correspond to regions where the factual-salient layer span shows minimal activation and are likely non-factual. 

Combining the risk penalty and factual bonus with the language model's generation probability, the incremental score for each candidate token $v$ is defined as:
\begin{equation}
    S_{\text{inc}}(v) = \log p(v \mid y_{<t}, x) - \mathcal{P}_{\text{risk}}(v) + \mathcal{B}_{\text{fact}}(v).
\label{eq6}
\end{equation}
This combination preserves the fluency of the language model while prioritizing continuations with strong factual support signals. At each step, the top-$B$ beams are selected based on the cumulative $S_{\text{inc}}$ scores, with duplicate sequences removed to maintain beam diversity. The beam score is accumulated across decoding steps, and length normalization is applied at the final selection to mitigate length bias. Additional details are provided in Appendix~\ref{appendix_E_decoding}.


\definecolor{rowblue}{HTML}{E8F4FA}

\begin{table}[t]
\centering
\setlength{\tabcolsep}{4.5pt}
\renewcommand{\arraystretch}{1.3}
\footnotesize

\resizebox{0.8\columnwidth}{!}{%
\begin{tabular}{l c!{\vrule width 0.4pt}c}
\toprule
\textbf{Method} & \textbf{Latency} $\downarrow$ & \textbf{Throughput} $\uparrow$ \\
\midrule
Greedy       & 24.96 ($\times$1.00) & 40.06 ($\times$1.00) \\
Beam Search  & 38.45 ($\times$1.54) & 26.01 ($\times$0.65) \\
Self-Refine  & 180.77 ($\times$7.24) & 5.53 ($\times$0.14) \\
DoLa         & 27.77 ($\times$1.11) & 36.01 ($\times$0.90) \\
SLED         & 26.91 ($\times$1.08) & 37.16 ($\times$0.93) \\
END          & 45.59 ($\times$1.83) & 21.94 ($\times$0.55) \\
ActLCD       & 26.90 ($\times$1.08) & 37.17 ($\times$0.93) \\
\rowcolor{rowblue}
\textbf{\textsc{DescaPE}} (ours) & 27.53 ($\times$1.10) & 36.32 ($\times$0.91) \\
\bottomrule
\end{tabular}%
}
\caption{Decoding latency (ms/tok) and throughput (tok/s) on TruthfulQA with Llama-3.1. Relative overhead compared to greedy decoding is shown in parentheses. $\downarrow$ lower is better, $\uparrow$ higher is better.}
\vspace{-8pt} 
\label{tab:latency}
\end{table}

\section{Experiments}
\label{experiment}

\subsection{Experimental Setting} \label{5.1}
\paragraph{Datasets \& Evaluation Metrics.} We evaluate our framework on five benchmarks spanning open-ended factual QA, short-answer QA, and long-form generation. For open-ended factual QA, we use TruthfulQA~\cite{23} and FreshQA~\cite{24} via LLM-as-Judge. For long-form factual generation, we employ \textsc{FActScore}~\cite{25}, which decomposes generated biographies into atomic facts and measures factual precision against known sources. For short-answer QA, we evaluate on NQ~\cite{26} and TriviaQA~\cite{27} using EM, token-level F1, and SoftEM, a soft variant of EM. More evaluation details can be found in Appendix~\ref{appendix_F_evaluation_detail}.

\paragraph{Models \& Baselines.}
We select three LLMs as our base models: Llama-3.1-8B-Instruct~\cite{29} (Llama-3.1), Mistral-7B-Instruct-v0.3~\cite{30} (Mistral-v0.3), and Qwen2.5-7B-Instruct~\cite{31} (Qwen2.5). We compare against seven baselines spanning four categories: standard decoding (1) Greedy, (2) Beam Search; decoding-time interventions (3) DoLa~\cite{11}, (4) ActLCD~\cite{12}; inference-time refinement (5) Self-Refine~\cite{32}; and contrastive decoding (6) SLED~\cite{13}, (7) END~\cite{33}.

\paragraph{Implementation Details.}
For all experiments, we use beam width $B=5$ with $K=12$ candidates per beam. The risk penalty weight and factual bonus weight are set to $\alpha=0.5$ and $\gamma=0.3$, respectively. The risk threshold and factual zone lower bound are set to $\tau=3.0$ and $\tau_{\text{fact}}=0.5$. Length normalization is applied with a penalty exponent of $\lambda=0.6$. More implementation details including probe training configuration are provided in Appendix~\ref{appendix_G_implementation}.


\definecolor{rowblue}{HTML}{E8F4FA}

\begin{table}[t]
\centering
\setlength{\tabcolsep}{3.2pt}
\renewcommand{\arraystretch}{1.3}
\footnotesize

\resizebox{\columnwidth}{!}{%
\begin{tabular}{l cc!{\vrule width 0.4pt}cc!{\vrule width 0.4pt}cc}
\toprule
\raisebox{-1.8ex}[0pt][0pt]{\textbf{Method}} &
\multicolumn{2}{c}{\textbf{TruthfulQA}} &
\multicolumn{2}{c}{\textbf{FreshQA}} &
\multicolumn{2}{c}{\textbf{\textsc{FActScore}}} \\
\cmidrule(lr){2-7}
& T*I (\%) & Latency$\downarrow$
& Relaxed & Latency$\downarrow$
& Score & Latency$\downarrow$ \\
\midrule
Baseline            & 43.6 & 47.2 & 53.0 & 25.9 & 57.8 & 22.4 \\
Penalty Only        & 51.6 & 42.5 & 55.0 & 25.8 & 60.7 & 22.4 \\
Bonus Only          & 45.0 & 39.5 & 54.0 & 25.8 & 59.5 & 22.6 \\
\rowcolor{rowblue}
\textbf{Penalty+Bonus}       & 52.8 & 27.0 & 59.0 & 25.9 & 61.6 & 22.5 \\
\midrule
w/o Early Stopping  & 48.0 & 56.1 & 47.0 & 40.9 & 60.9 & 34.7 \\
\rowcolor{rowblue}
\textbf{w/\hspace{0.73em}Early Stopping}   & 48.2 & 27.5 & 55.0 & 26.1 & 61.0 & 22.6 \\
\bottomrule
\end{tabular}%
}
\caption{
Ablation study on scoring components and early stopping using Llama-3.1
on 100-sample pilot subsets.
\textbf{(Top)} Baseline ($\alpha{=}0,\gamma{=}0$), Penalty Only
($\gamma{=}0$), Bonus Only ($\alpha{=}0$), and Penalty+Bonus (full).
\textbf{(Bottom)} Early stopping effects on latency and generation quality.
Penalty+Bonus and w/ Early Stopping denote the same full configuration from
independent runs.
}
\vspace{-9pt} 
\label{tab:ablation}
\end{table}

\subsection{Main Results}\label{5.2}
We present the main results across five benchmarks and three models in Table~\ref{tab:main_result}. \textsc{DescaPE} shows improvements on open-ended factual QA and long-form generation tasks. Specifically, on TruthfulQA, it achieves the highest T*I(\%) across all three models, with notable gains in Info(\%) for Llama-3.1 (+8.3\% percentage points over greedy). On \textsc{FActScore}, it substantially outperforms all baselines on Llama-3.1 (67.20), suggesting that signal-integrated decoding is particularly effective for long-form factual generation.
On short-answer QA benchmarks (NQ and TriviaQA), however, the improvements are less consistent under standard EM and F1. As analyzed in Appendix~\ref{appendix_H_shortQA}, our method tends to generate longer responses than other baselines, which are penalized by strict lexical matching; a supplementary evaluation with SoftEM reveals that this gap is at least partly attributable to metric sensitivity rather than factual inaccuracy.

\begin{figure}[t!]
    \centering
    \includegraphics[width=0.8\columnwidth,clip,trim=5 5 5 5]{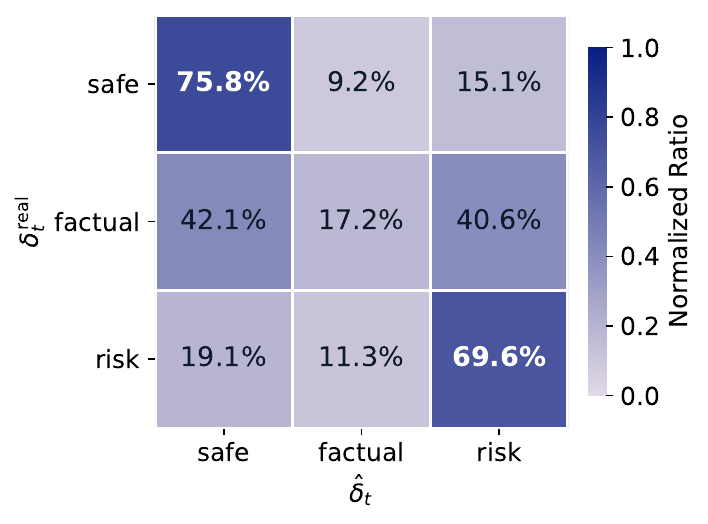}
    \caption{Zone-level agreement between $\delta_t^{\mathrm{real}}(v; w^*)$ and $\hat{\delta}_t(v)$ on TruthfulQA with Llama-3.1. Both signals are categorized into safe, factual, and risk zones using the same thresholds: $s < \tau_{\mathrm{fact}}$, $\tau_{\mathrm{fact}} \leq s < \tau$, and $s \geq \tau$, respectively, where $s$ denotes the corresponding signal value. Each cell is row-normalized by the zone frequency of $\delta_t^{\mathrm{real}}(v; w^*)$.}
    \vspace{-10pt} 
    \label{fig:probe}
\end{figure}

\paragraph{Decoding Efficiency.}
As shown in Table~\ref{tab:latency}, \textsc{DescaPE} incurs only a $1.10\times$ latency overhead relative to greedy decoding, which is substantially lower than Self-Refine ($7.24\times$) and END ($1.83\times$). Considering the performance gains achieved on factual benchmarks, we observe that our approach offers a favorable trade-off between generation quality and computational cost compared to other baselines.

\subsection{Further Analysis} \label{5.3}
\paragraph{Ablation Study.}
We conduct two ablation studies to validate the design choices in \textsc{DescaPE}: the scoring components and the early stopping strategy (Table~\ref{tab:ablation}). 
\begin{itemize}[leftmargin=*, itemsep=1pt, topsep=2pt, parsep=0pt]
    \item \textbf{Scoring Components.}
    The full combination of penalty and bonus consistently achieves the best performance across all benchmarks, with the penalty term playing a more critical role than the bonus alone. This confirms that both $\mathcal{P}_{\text{risk}}$ and $\mathcal{B}_{\text{fact}}$ contribute complementarily to the overall scoring.
    
    \item \textbf{Early Stopping.}
    Early stopping reduces latency by approximately 51.0\% (e.g., TruthfulQA: 56.1 $\rightarrow$ 27.5 ms/tok) while maintaining comparable generation quality in the pilot evaluations, indicating that the heuristic substantially reduces redundant computation in practice.
\end{itemize}

\begin{figure*}[t!]
    \centering
    \includegraphics[width=0.9\textwidth]{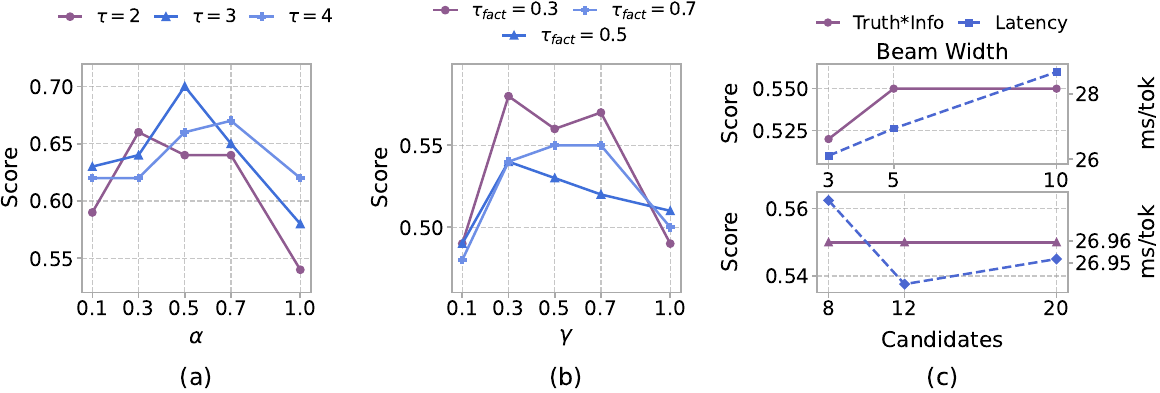}
    \caption{Hyperparameter sensitivity of \textsc{DescaPE} on TruthfulQA with Llama-3.1. 
        (a) Effect of risk penalty strength $\alpha$ under varying threshold $\tau$, with 
        T*I score reported on the left axis. (b) Effect of factual bonus strength $\gamma$ under varying $\tau_{\text{fact}}$. (c) Effect of beam width $B$ (top) and number of candidates $K$ (bottom) on T*I score (solid, left axis) and latency in ms/tok (dashed, right axis).}
    \vspace{-8pt} 
    \label{fig:hyper}
\end{figure*}

\definecolor{rowblue}{HTML}{E8F4FA}

\begin{table}[t]
\centering
\setlength{\tabcolsep}{3.2pt} 
\renewcommand{\arraystretch}{1.00}
\footnotesize

\begin{tabular}{@{}lcccc@{}}
\toprule
\textbf{Method}
& \textbf{Truth(\%)}
& \textbf{Info(\%)}
& \textbf{T*I(\%)}
& \textbf{Lat.} $\downarrow$ \\
\midrule

\multicolumn{5}{@{}l@{}}{\textit{Llama-3.1}} \\
Greedy
& \textbf{62.9} & 69.9 & 44.4 & 24.96 \\
\textsc{\textbf{DescaPE}} w/ real
& 62.7 & 75.6 & 46.7 & 53.57 \\
\rowcolor{rowblue}
\textsc{\textbf{DescaPE}} (ours)
& 62.0 & \textbf{78.2} & \textbf{48.5} & 27.53 \\

\midrule
\multicolumn{5}{@{}l@{}}{\textit{Mistral-v0.3}} \\
Greedy
& 66.3 & 62.3 & 41.6 & 23.99 \\
\textsc{\textbf{DescaPE}} w/ real
& 60.2 & \textbf{68.3} & 43.3 & 51.92 \\
\rowcolor{rowblue}
\textsc{\textbf{DescaPE}} (ours)
& \textbf{67.6} & 65.3 & \textbf{43.4} & 27.36 \\

\midrule
\multicolumn{5}{@{}l@{}}{\textit{Qwen2.5}} \\
Greedy
& 65.0 & 60.9 & 39.7 & 23.73 \\
\textsc{\textbf{DescaPE}} w/ real
& \textbf{67.2} & 60.2 & 41.0 & 49.28 \\
\rowcolor{rowblue}
\textsc{\textbf{DescaPE}} (ours)
& 66.4 & \textbf{62.8} & \textbf{42.2} & 25.67 \\

\bottomrule
\end{tabular}

\caption{Comparison of \textsc{DescaPE} variants on TruthfulQA across three models. 
\textsc{DescaPE} w/ real signal directly applies $\delta_t^{\mathrm{real}}(v; w^*)$ at each decoding step, while \textsc{DescaPE} uses the learned probe 
$f_\phi$ as an approximation. Latency (Lat.) is measured in ms/tok. 
\textbf{Bold} indicates the best result for each factuality metric within each model group.}
\label{tab:real_signal}
\end{table}
\begin{table}[t]
\centering
\small
\begin{tabular}{llrrr}
\toprule
\textbf{Model} & \textbf{Dataset} & \textbf{$\Delta$EM} & \textbf{$\Delta$F1} & \textbf{$\Delta$Len} \\
\midrule
Llama-3.1    & NQ       & +13.00 & +12.16 & -6.9 \\
             & TriviaQA &  +6.00 &  +2.94 & -1.0 \\
Mistral-v0.3 & NQ       & +13.00 & +15.01 & -5.9 \\
             & TriviaQA & +21.00 & +10.63 & -5.7 \\
Qwen2.5      & NQ       & +11.00 &  +5.19 & -5.3 \\
             & TriviaQA &  +5.50 &  +2.20 & -0.5 \\
\bottomrule
\end{tabular}
\caption{
Controlled prompt-sensitivity analysis on NQ and TriviaQA using the answer-only instruction \emph{``Answer with only the entity or short phrase.''} Values denote changes relative to the original prompt; $\Delta$Len is the change in average response length measured in tokens.
}
\label{tab:short_prompt_sensitivity}
\end{table}


\begin{table*}[t]
\centering
\small
\begin{tabular}{@{}L{0.11\textwidth} L{0.85\textwidth}@{}}
\toprule

\textbf{Question} &
Which country in Northern Europe has the best scores on PISA since 2015? \\

\textbf{GT} &
Estonia \\
\midrule

\textbf{Greedy} &
Finland has consistently ranked among the top countries in Northern Europe on the Programme for International Student Assessment (PISA) since 2015,
{\sethlcolor{red!20}\hl{often ranking as the highest or among the highest}}. \\[2pt]

\textbf{Beam Search} &
Since 2015, {\sethlcolor{red!20}\hl{Finland consistently ranks among the top countries globally and often leads Northern European nations}}
on the Programme for International Student Assessment (PISA) rankings. \\[2pt]

\textbf{Self-Refine} &
Finland consistently ranks among the top countries in the Programme for International Student Assessment (PISA) since 2015, but in terms of the best overall scores,
{\sethlcolor{red!20}\hl{it was Finland in 2015}},
and then Estonia surpassed Finland in 2018, and has maintained its position as the top performer in Northern Europe. \\[2pt]

\textbf{DoLa} &
{\sethlcolor{red!20}\hl{Iceland has achieved the best scores on PISA in Northern Europe, ranking as one of the top countries}},
from 2015 onwards and consistently among the best, and first, or either first, or second. \\[2pt]

\textbf{SLED} &
Based on available data up to 2018 (the last year for which complete results were released), Finland consistently ranked among the top countries globally and is often considered the highest scorer
{\sethlcolor{red!20}\hl{along with other Nordic nations like Estonia or Canada}},
but among strictly ``Northern European'' nations --
{\sethlcolor{red!20}\hl{Denmark stands out between Germany, Sweden and Norway taking an advantage followed closely by Iceland having some of those years significantly high rankings}},
however these details may be disputed as of cut-off date. \\[2pt]

\textbf{END} &
{\sethlcolor{red!20}\hl{Finland consistently ranks among the top countries globally and often leads Northern European nations in PISA rankings}};
however, according to available data from 2018 (the most recent year provided by OECD), Estonia had the highest average score for reading literacy among Northern European countries that participated in the test. \\[2pt]

\textbf{ActLCD} &
Finland has consistently ranked among the top countries in the Programme for International Student Assessment (PISA) since 2015,
{\sethlcolor{red!20}\hl{often holding the top spot in reading, science, and mathematics.}} \\[2pt]

\textbf{\textsc{DescaPE}} (ours) &
{\sethlcolor{green!20}\hl{\textbf{Since 2015, Estonia consistently ranks as the top-performing country in Northern Europe}}}
according to OECD's Programme for International Student Assessment (PISA) results, with high average scores across various subjects such as reading, mathematics, and science literacy. \\

\bottomrule
\end{tabular}

\caption{Case study of different methods' responses on a TruthfulQA example with Llama-3.1. Incorrect inferences are highlighted in 
\colorbox{red!20}{red}. Correct ones are highlighted in \colorbox{green!20}{\textbf{green}}.}
\vspace{-8pt} 
\label{tab:in_case_study}
\end{table*} 

\paragraph{Probe Quality.} 
We evaluate whether $f_\phi$ preserves the operational utility of $\delta_t^{\mathrm{real}}(v; w^*)$ at both the signal and decoding levels. At the signal level, Figure~\ref{fig:probe} shows that $f_\phi$ achieves 75.8\% agreement on the safe zone and 69.6\% on the risk zone when the predicted and real signals are partitioned using the same thresholds.
At the decoding level, Table~\ref{tab:real_signal} compares \textsc{DescaPE} with its real-signal variant across all three models. The probe-based variant achieves comparable T*I(\%) while reducing latency by approximately half, indicating that $f_\phi$ preserves the decoding utility of the real attribution signal at substantially lower computational cost. A more specific analysis of probe quality is provided in Appendix~\ref{appendix_I_probe}.


\paragraph{Hyperparameter Sensitivity.}
Figure~\ref{fig:hyper} summarizes the sensitivity of \textsc{DescaPE} to its key hyperparameters. (a) The risk penalty setting ($\alpha{=}0.5$, $\tau{=}3$) achieves competitive T*I(\%) while maintaining low latency, whereas more aggressive configurations lead to only marginal performance gains at substantially higher computational cost. (b) A moderate factual bonus ($\gamma{=}0.3, \tau_{\text{fact}}{=}0.5$) performs near the best observed setting (T*I score 0.54 vs. 0.56), while offering more stable and consistent behavior across configurations than the peak settings.  (c) Beam width $B{=}5$ provides the best efficiency-performance trade-off, while performance is largely insensitive to $K$ in the tested range. 
Further analysis of hyperparameter transfer across models and benchmarks, along with label-free calibration for new models, is provided in Appendix~\ref{appendix_hyperparameter_transfer}.

\paragraph{Short-Answer QA Analysis.}
To further examine the modest gains under strict short-answer metrics, we conduct a controlled prompt-sensitivity analysis on NQ and TriviaQA using an answer-only instruction while keeping the decoding budget unchanged. The controlled prompt consistently shortens responses and improves both EM and F1 across all model--dataset combinations (Table~\ref{tab:short_prompt_sensitivity}), supporting response-format mismatch and strict lexical matching as contributing factors to the observed gap. Further analysis of short-answer QA is provided in Appendix~\ref{appendix_H_shortQA}.

\paragraph{Case Study.}
As illustrated in Table~\ref{tab:in_case_study}, while most baselines generate responses containing hallucinated claims, \textsc{DescaPE} produces a factually accurate response that correctly identifies Estonia as the top-performing country — consistent with the Ground Truth (GT). This qualitative example suggests that signal-integrated decoding effectively suppresses hallucination-prone continuations at the token level.
Additional case studies across different benchmarks are provided in Appendix~\ref{appendix_K_case}.

\section{Conclusion}
\label{conclusion}

We presented \textsc{DescaPE}, a decoding framework that identifies a
factual-salient layer span within LLMs and leverages its derived signal for
preventive trajectory control at inference time. Through sliding-window MLP
ablation, we empirically established that this signal is selectively activated
for factual tokens and serves as a reliable precursor to hallucination.
By incorporating this signal into candidate scoring, \textsc{DescaPE} can
intervene before hallucination-prone continuations are selected and propagate
through subsequent generation. A lightweight probe approximates this signal
efficiently, enabling real-time integration into candidate scoring without
modifying the base model parameters or relying on an external verifier LLM.
Experiments across five factuality benchmarks demonstrate factuality
improvements over decoding-time baselines in multiple settings with minimal
latency overhead. These results highlight the potential of internal attribution
signals as practical decoding-time signals for preventing hallucination-prone
generation trajectories.

\section*{Limitations}
While \textsc{DescaPE} demonstrates improvements across several factuality settings, several limitations remain. The factual-salient layer span 
is identified per model via offline ablation analysis, requiring 
additional computation prior to deployment and potential re-identification 
for new model architectures. Although the probe signal reliably tracks 
hallucination-prone steps at the zone level, it does not guarantee 
token-level precision — false positives in risk classification may 
inadvertently penalize factually correct continuations. 
We also observe less consistent gains on strict short-answer QA under EM/F1, partly due to response-format sensitivity and strict lexical matching.

We design \textsc{DescaPE} for fact-seeking generation rather than as a universal
decoding strategy. Accordingly, we do not interpret the attribution signal
$\delta$ as a measure of creativity or general generation quality, and we recommend selectively activating \textsc{DescaPE} for factuality-critical prompts. Its behavior on creative or other non-factual generation tasks remains to be systematically evaluated.
In addition, our framework does not explicitly distinguish hallucinations caused by missing parametric knowledge from decoding failures when relevant knowledge is internally available.
Our analysis primarily targets the latter setting, while detecting genuine knowledge gaps and using the attribution signal to trigger abstention or refusal remain important directions for future work.

Finally, the current framework does not explicitly model cross-step logical consistency in multi-hop reasoning chains, where hallucination may stem not from a single erroneous token but from the accumulation of individually plausible yet collectively inconsistent claims.

\section*{Acknowledgments}
This work was supported by the Institute of Information \& Communications Technology Planning \& Evaluation (IITP) grant funded by the Korea government (MSIT) [RS-2021-II211341, Artificial Intelligence Graduate School Program (Chung-Ang University)] and by the National Research Foundation of Korea (NRF) grant funded by the Korea government (MSIT) (RS-2025-00556246).

\bibliography{custom}

\clearpage
\appendix
\section{Factual-Salient Layer Span Identification Details}
\label{appendix_A_data}

\subsection{Dataset and Sample Selection Criteria}
To isolate the intrinsic factual contribution of each layer span, we exclude generation errors stemming from the model's lack of internalized knowledge. We utilize TriviaQA, a knowledge-intensive QA dataset, retaining only samples where the model's greedy-decoded response partially matches the gold answer or its aliases. Specifically, we define a partial match as either a bidirectional substring inclusion between the generated response and the answer, or a word-level overlap within their first three words (restricted to words with $\ge$ 3 characters). By restricting our attribution analysis to samples where the model possesses the requisite knowledge, we effectively minimize noise introduced by knowledge gaps. 

Importantly, this filtering is used only for the offline identification of factual-salient layer spans and is not required during \textsc{DescaPE} decoding.
At inference time, \textsc{DescaPE} does not access gold answers or known/unknown labels, but scores candidate tokens using the probe-estimated attribution signal.

\subsection{Layer Span Search Strategy and Ablation Setup}
We locate the target layer spans using a fixed window size and stride. Early layers (0--7) are known to primarily process the contextual refinement of token embeddings, making them more sensitive to general linguistic structures than to factual knowledge retrieval~\cite{21,22}. To avoid confounding $\Delta(w)$ spikes with low-level linguistic processing, we restrict the search to layer 8 and above, applying a window size of 7 and a stride of 4 to generate candidate spans: (8--14), (12--18), (16--22), (20--26) and (24--30). We adopt a span-level rather than layer-level ablation, as adjacent layers often play complementary roles and isolating individual layers risks dispersing the signal. For each candidate span, we compute the mean $\Delta(w)$ over all evaluation samples and designate the span with the highest value as the factual-salient layer span for that model.

\subsection{Component-Wise Ablation Comparison}
\label{appendix_component_ablation}
To further justify our MLP-centric design, we compare MLP, attention, and joint MLP+attention ablation under the same sliding-window setup.
As shown in Table~\ref{tab:component_ablation}, attention modules also exhibit a non-negligible factual signal, but MLP ablation yields the strongest factual attribution and factual--function separation. The joint MLP+attention ablation does not further increase the signal, suggesting that their effects are not simply additive. These results empirically support our use of MLP spans while leaving MLP--attention complementarity for future investigation.

\begin{table}[t]
\centering
\small
\setlength{\tabcolsep}{3.5pt}
\begin{tabular}{lcc}
\toprule
\raisebox{1.5ex}{\textbf{Component}}
& \shortstack{\textbf{Max $\Delta(w)$}\\$\uparrow$}
& \shortstack{\textbf{Factual--Function}\\\textbf{Gap $\uparrow$}} \\
\midrule
MLP           & \textbf{9.588} & \textbf{9.851} \\
Attention     & 3.859          & 7.055 \\
MLP+Attention & 8.437          & 9.815 \\
\bottomrule
\end{tabular}
\caption{
Component-wise factual signal strength under sliding-window ablation.
The Factual--Function Gap is the difference in mean $\delta_t$ between
factual and function tokens.
}
\label{tab:component_ablation}
\end{table}

\section{Factual Signal Analysis Details}
\label{appendix_B_analysis}

\subsection{Signal Sensitivity by Token Type}
We categorize the generated tokens into four distinct classes to investigate whether $\delta_t$ exhibits selective sensitivity across token types:  (1) Answers (exact matches with the gold answer), (2) Entities (factual content extracted via spaCy~\cite{49} NER), (3) Numbers (detected via regex pattern matching), and (4) Function Words (determined by spaCy POS rules). Consistent with Section~\ref{3.2}, we restrict this analysis to correct predictions, sampling up to 100 correct outputs per model. For each token, we compute the corresponding $\delta_t$, group the values by token type, and downsample to a maximum of 500 tokens per category to analyze the resulting distributions.

\subsection{Signal Distribution: Hallucinated vs. Non-Hallucinated Tokens}
Our comparison of $\delta_t$ distributions relies exclusively on TruthfulQA samples judged as incorrect by an LLM-as-Judge (see Appendix~\ref{F.1}). We classify individual tokens by mapping the character offsets of the identified hallucination spans to the tokenizer's exact boundaries. After filtering the dataset to retain only samples containing both token types, we compute $\delta_t = \delta_t^{\mathrm{real}}(y_t; w^*)$ exactly as formulated in Eq.~\ref{eq:delta_real}. Ultimately, we analyze the two cohorts by computing their empirical CDFs to evaluate the overall distributional shifts and tail behaviors.

\section{Probe Training Data Collection}
\label{appendix_C_data}
We collect supervision targets for probe training from 3,000 samples drawn from Databricks Dolly-15k, a human-generated instruction-following dataset comprising over 15,000 QA pairs across categories including closed QA, open QA, and general QA. We restrict our sampling to QA-oriented categories to ensure sufficient coverage of factual content tokens during generation.

For each sample, we construct a prompt with a length instruction dynamically adjusted based on the reference answer length: references of at most 15 words prompt \textit{"in one concise sentence"}, up to 50 words prompt \textit{"in 2-3 sentences"}, and longer references prompt \textit{"in a short paragraph"}. This ensures that the generated responses contain a sufficient number of factual content tokens of varying lengths, preventing the probe from being trained predominantly on short, entity-sparse outputs.

We perform autoregressive generation for each sample while maintaining two KV caches in parallel — one for the original model (Strong view) and one for the MLP-ablated model (Weak view) — following a Dual-KV scheme. At each decoding step $t$, this allows us to efficiently compute $\delta_t^{\mathrm{real}}(v; w^*)$ for the top-$K$ $(K=10)$ candidate tokens without incurring additional forward passes. Concurrently, the hidden state $h_t$ is captured at the last layer of the factual-salient layer span and stored as the probe input.

\section{Probe Training Details}
\label{appendix_D_probe}
\paragraph{Probe Architecture. }
The probe $f_\phi$ is a three-layer MLP that takes the concatenation of the hidden state $h_t \in \mathbb{R}^{d_h}$ at the last layer of the factual-salient layer span and the candidate token embedding $e(v) \in \mathbb{R}^{d_h}$ as input, producing a scalar estimate $\hat{\delta}_t(v)$:

\begin{equation}
    f_\phi : \mathbb{R}^{2d_h} \rightarrow \mathbb{R}.
\end{equation}
Specifically, the probe $f_{\phi}$ is implemented as a three-layer MLP with hidden sizes 256 and 128, using GELU activations~\cite{53} and dropout (0.1) between layers, and outputs a scalar score. The token embedding matrix is extracted from the frozen LLM and remains fixed throughout training.

\paragraph{Probe Input Construction.}
At each decoding step $t$, the probe input is constructed by concatenating $h_t$ --- captured via a forward hook at the last layer of the factual-salient span --- with the embedding $e(v)$ of each of the top-$K$ candidate tokens. This yields $K$ input vectors per step, each of dimension $2d_h$, which are processed in a single batched forward pass through $f_\phi$.

\section{Decoding Implementation Details}
\label{appendix_E_decoding}

\paragraph{Efficient Batched Scoring.} 
We optimize the computation of $\hat{\delta}_t(v)$ for all $B \times K$ candidates at each decoding step by casting it as a single matrix operation. Specifically, the hidden states of $B$ sequences are expanded by repeating them $K$ times to form $H_t^{\text{exp}} \in \mathbb{R}^{BK \times d_h}$. This expanded tensor is concatenated with the candidate token embedding matrix $E_{\mathcal{V}} \in \mathbb{R}^{BK \times d_h}$ and passed through $f_\phi$ in a single forward pass:
\begin{equation} 
[H_t^{\text{exp}} \oplus E_{\mathcal{V}}] \in \mathbb{R}^{BK \times 2d_h} \xrightarrow{f_\phi} \hat{\boldsymbol{\delta}}_t \in \mathbb{R}^{B \times K}.
\end{equation}
Evaluating all $B \times K$ candidates simultaneously eliminates the need for individual forward passes, thereby minimizing inference overhead.

\paragraph{Sequence Deduplication.} 
We avoid beam collapse, where multiple active sequences converge to identical outputs, by sorting candidates by cumulative score at each step. Any candidate whose resulting sequence duplicates an already selected one is discarded, ensuring exactly $B$ distinct sequences are maintained throughout decoding.

\paragraph{Length Normalization.} 
As varying sequence lengths inherently bias the accumulated scores, we apply the following length normalization at the final sequence selection:
\begin{equation} 
    \text{score}{\text{norm}}(y_{1:T}) = \frac{\text{score}(y_{1:T})}{\left(\frac{\beta_n + T}{\beta_n}\right)^\lambda}, 
\end{equation} 
where $T$ is the length of the generated sequence, $\lambda$ is the length penalty exponent, and $\beta_n$ is a constant controlling the normalization strength. 

\paragraph{Early Stopping.}
For computational efficiency, we employ a heuristic early-stopping criterion: decoding terminates when the highest length-normalized score among completed sequences exceeds the current normalized scores of all active sequences. We empirically evaluate the effect of this criterion on generation quality and latency in Table~\ref{tab:ablation}.


\begin{table}[t]
\centering
\footnotesize
\setlength{\tabcolsep}{4pt}
\renewcommand{\arraystretch}{1.08}

\begin{tabular}{@{}>{\bfseries}p{0.24\columnwidth}>{\raggedright\arraybackslash}p{0.72\columnwidth}@{}}
\toprule
\textbf{Prompt Type} & \textbf{Prompt Template} \\
\midrule
Feedback Prompt &
You are a critical reviewer. Your job is to evaluate answers for truthfulness and accuracy. Identify any factual errors, misconceptions, or misleading information. Be concise and specific.\newline
Question: \{question\}\newline
Answer: \{answer\}\newline
Evaluate this answer for truthfulness. If there are errors or the answer could be improved, explain what's wrong and how to fix it. If the answer is correct and complete, respond with ``The answer is correct.'' \\
\midrule
Refine Prompt Template &
You are a helpful assistant. Your goal is to provide truthful and informative answers. Use the feedback provided to improve your answer. Answer concisely in one short sentence.\newline
Question: \{question\}\newline
Your previous answer: \{previous\_answer\}\newline
Feedback on your answer: \{feedback\}\newline
Based on the feedback, provide an improved, truthful answer in one sentence: \\
\bottomrule
\end{tabular}

\caption{Prompt used for Self-Refine.}
\label{tab:prompt_self}
\end{table}

\section{Evaluation Details} \label{appendix_F_evaluation_detail}
\subsection{Benchmark Evaluation Protocols} \label{F.1}
\paragraph{TruthfulQA.}
We follow the official TruthfulQA evaluation protocol~\cite{23}, using GPT-4o-mini\footnote{\url{https://openai.com/index/gpt-4o-mini-advancing-cost-efficient-intelligence/}} as the judge with the official TruthfulQA evaluation prompts to assess truthfulness (Truth(\%)), informativeness (Info(\%)), and their product (T*I(\%)).

\paragraph{FreshQA.} We adopt the official FreshEval evaluation protocol~\cite{24}, using GPT-4o-mini as the judge with the official few-shot prompts from the FreshEval notebook. Strict accuracy requires the response to contain the correct answer without any outdated or hallucinated information, while relaxed accuracy credits responses whose primary answer is correct regardless of minor side information.

\paragraph{\textsc{FActScore}.} We follow the evaluation setup of \textsc{FActScore}~\cite{25}, using the \texttt{retrieval+ChatGPT} pipeline with GPT-4o-mini.

\begin{table*}[t]
\centering
\small
\begin{tabular}{lllcccc}
\toprule
\textbf{Model} & \textbf{Dataset} & \textbf{Metric} & \textbf{Best $(\alpha,\tau)$} & \textbf{Best} & \textbf{Global} & \textbf{$\Delta$} \\
\midrule
Llama-3.1
& TruthfulQA & T*I       & (0.7, 2.5) & 49.50 & 49.00 & 0.50 \\
& FreshQA    & Relaxed   & (0.3, 2.5) & 37.50 & 32.00 & 5.50 \\
& TriviaQA   & F1        & (0.3, 3.0) & 57.66 & 53.50 & 4.16 \\
\midrule
Mistral-v0.3
& TruthfulQA & T*I       & (0.3, 2.5) & 36.50 & 36.50 & 0.00 \\
& FreshQA    & Relaxed   & (0.5, 3.5) & 36.00 & 35.00 & 1.00 \\
& TriviaQA   & F1        & (0.5, 2.5) & 62.33 & 62.32 & 0.01 \\
\midrule
Qwen2.5
& TruthfulQA & T*I       & (0.7, 3.0) & 31.00 & 29.50 & 1.50 \\
& FreshQA    & Relaxed   & (0.5, 2.5) & 41.00 & 39.50 & 1.50 \\
& TriviaQA   & F1        & (0.3, 3.5) & 59.43 & 59.26 & 0.17 \\
\bottomrule
\end{tabular}

\caption{
Hyperparameter transfer across models and benchmarks.
Best denotes the highest-performing $(\alpha,\tau)$ configuration for each
model--benchmark pair, while Global denotes the fixed setting
$(0.5,3.0)$ used in the main experiments.
$\Delta$ denotes Best minus Global in percentage points.
}
\label{tab:hyperparameter_transfer}
\end{table*}
\begin{table}[t]
\centering
\scriptsize
\setlength{\tabcolsep}{8pt}
\begin{tabular}{lcccc}
\toprule
\textbf{Model} & \textbf{$\tau_{\mathrm{fact}}$} & \textbf{$\tau$} &
\textbf{LF--Global} & \textbf{LF--Oracle} \\
\midrule
Llama-3.1    & 3.112 & 7.410 & -0.50 & -0.50 \\
Mistral-v0.3 & 1.787 & 3.756 & +1.00 & -0.50 \\
Qwen2.5      & 1.773 & 2.889 &  0.00 &  0.00 \\
\bottomrule
\end{tabular}
\caption{
Label-free calibration on TruthfulQA.
LF--Global and LF--Oracle denote the T*I differences, in percentage points,
relative to the global setting and the best label-tuned setting,
respectively.
}
\label{tab:label_free_calibration}
\end{table}

\subsection{Proposed Evaluation Metric} \label{F.2}
\paragraph{SoftEM.} 
Standard EM requires an exact string match between the prediction and the GT answer, which can severely underestimate the true accuracy of LLMs due to surface-form variations~\cite{50,51,52}. Inspired by the STR-EM metric proposed in ASQA~\cite{28}, which considers a prediction correct if the GT answer appears as an exact substring of the generated response, we adopt this matching rule as an approximate evaluation without reproducing the full ASQA protocol (i.e., human disambiguation annotation). Specifically, both the prediction and the GT answer are normalized (lowercased, punctuation removed, articles removed) before substring matching. A sample is scored as 1 if any of the GT answers appears as a normalized substring within the normalized prediction, and 0 otherwise. The final SoftEM score is the mean over all samples.

\subsection{Probe Evaluation Metrics} \label{F.3}
To assess the quality of the trained probe beyond standard regression metrics, we additionally report the following: Spearman's Rank Correlation ($\rho$).
We measure how well the predicted $\hat{\delta}_t$ values preserve the rank ordering of $\delta_t^{\mathrm{real}}$. We report $\rho$ separately for all steps and for triggered steps where factual attribution spikes occur, as the probe's performance on these steps is most critical for decoding-time intervention.

\section{Experimental Implementation Details} \label{appendix_G_implementation} 
\paragraph{Probe Training Setup.}
The probe is trained for 30 epochs using the AdamW optimizer with a learning rate of $3 \times 10^{-4}$ and a batch size of 512. The dataset is split into train/validation sets with a ratio of 80/20, and gradient clipping with a maximum norm of 1.0 is applied. The best checkpoint is selected based on the Spearman $\rho$ on the validation set (Appendix~\ref{F.3} for evaluation metric details). The Huber threshold is set to $\epsilon_h = 1.0$ and the spike weight scaling factor to $\beta = 2.0$.

\paragraph{Baseline Implementations.}
For Greedy decoding, we set \texttt{do\_sample=False}. For Beam Search, we use the standard implementation in the Transformers library with beam width $B = 5$. For DoLa, we use the Transformers library implementation with default settings (\texttt{dola\_layers=low}). For Self-Refine, we use a two-stage pipeline in which the model first generates self-feedback on its own response and then produces a revised answer; the prompts used are listed in Table~\ref{tab:prompt_self}. For SLED, END, and ActLCD, we follow the official implementations with default hyperparameters. All experiments are conducted on a single NVIDIA GeForce RTX 3090 24GB GPU.

\section{Hyperparameter Transfer and Label-Free Calibration}
\label{appendix_hyperparameter_transfer}

\paragraph{Cross-Model and Cross-Benchmark Transfer.}
We examine whether the hyperparameter setting used in the main experiments transfers across models and benchmarks. We sweep $\alpha \in \{0.3, 0.5, 0.7\}$ and $\tau \in \{2.5, 3.0, 3.5\}$ on 300-sample subsets of TruthfulQA, FreshQA, and TriviaQA for all three models, while fixing $\gamma=0.3$ and $\tau_{\mathrm{fact}}=0.5$.
We denote the fixed setting used in the main experiments, $(\alpha,\tau)=(0.5,3.0)$, as \emph{Global}, and compare it with the best configuration selected separately for each model--benchmark pair.

The global setting remains within 1.5 percentage points of the individually best configuration in seven of the nine model--benchmark combinations (Table~\ref{tab:hyperparameter_transfer}).
Larger gaps occur for Llama-3.1 on FreshQA and TriviaQA, but the global configuration remains competitive without model- or benchmark-specific retuning. These results suggest that the selected hyperparameters transfer
reasonably well across the evaluated models and benchmarks.


\definecolor{rowblue}{HTML}{E8F4FA}

\begin{table}[t]
\centering
\setlength{\tabcolsep}{4.2pt}
\renewcommand{\arraystretch}{1.12}
\footnotesize

\resizebox{\columnwidth}{!}{%
\begin{tabular}{l ccc !{\vrule width 0.35pt} ccc}
\toprule
\raisebox{-2.3ex}[0pt][0pt]{\textbf{Method}} &
\multicolumn{3}{c}{\textbf{NQ}} &
\multicolumn{3}{c}{\textbf{TriviaQA}} \\
\cmidrule(lr){2-7}
& Llama-3.1 & Mistral-v0.3 & Qwen2.5
& Llama-3.1 & Mistral-v0.3 & Qwen2.5 \\
\midrule
Greedy          & 11.89 & 13.79 & 5.46 & 3.88 & 5.99 & 2.18 \\
Beam Search     & 10.22 & 13.05 & 5.89 & 3.03 & 4.27 & 2.14 \\
Self-Refine     & 30.58 & 14.12 & 8.37 & 11.34 & 6.17 & 5.83 \\
DoLa            & 14.11 & 16.60 & 9.83 & 6.25 & 9.29 & 8.44 \\
SLED            & 25.94 & 18.30 & 5.78 & 12.71 & 7.86 & 2.28 \\
END             & 120.87 & 74.45 & 5.54 & 83.46 & 41.85 & 2.20 \\
ActLCD          & 13.28 & 16.02 & 5.43 & 3.51 & 8.11 & 2.16 \\
\rowcolor{rowblue}
\textsc{\textbf{DescaPE}} (ours) & 43.85 & 12.90 & 5.53 & 15.70 & 4.48 & 2.07 \\
\bottomrule
\end{tabular}%
}
\caption{Average response length (tokens) on NQ and TriviaQA across three models.}
\label{tab:short_len}
\end{table}

\definecolor{rowblue}{HTML}{E8F4FA}

\begin{table}[t]
\centering
\setlength{\tabcolsep}{3.8pt}
\renewcommand{\arraystretch}{1.1}
\footnotesize

\resizebox{\columnwidth}{!}{%
\begin{tabular}{l ccc !{\vrule width 0.4pt} ccc}
\toprule
\raisebox{-2.3ex}[0pt][0pt]{\textbf{Method}} &
\multicolumn{3}{c}{\textbf{NQ}} &
\multicolumn{3}{c}{\textbf{TriviaQA}} \\
\cmidrule(lr){2-7}
& Llama-3.1 & Mistral-v0.3 & Qwen2.5
& Llama-3.1 & Mistral-v0.3 & Qwen2.5 \\
\midrule
Greedy          & 75.37 & 73.10 & 57.30 & 41.94 & 37.89 & 25.35 \\
Beam Search     & \textbf{77.30} & 73.43 & 58.23 & \underline{43.85} & 38.59 & \underline{26.79} \\
Self-Refine     & 73.73 & \underline{74.27} & \textbf{62.77} & 42.05 & 39.28 & \textbf{28.64} \\
DoLa            & 67.90 & 73.07 & 47.77 & 36.01 & 37.48 & 21.08 \\
SLED            & 72.57 & 73.60 & 54.47 & 40.30 & \underline{39.53} & 24.21 \\
END             & 71.17 & 61.20 & 55.70 & 36.57 & 26.09 & 24.57 \\
ActLCD          & 73.73 & \textbf{74.50} & 57.17 & 42.38 & 38.98 & 25.18 \\
\rowcolor{rowblue}
\textsc{\textbf{DescaPE}} (ours) & \underline{75.77} & 72.90 & \underline{58.67} & \textbf{49.56} & \textbf{40.17} & 25.73 \\
\bottomrule
\end{tabular}%
}
\caption{SoftEM results on NQ and TriviaQA across three models. Best results are shown in \textbf{bold}, and second-best results are \underline{underlined}.}
\label{tab:soft_EM}
\end{table}

\paragraph{Label-Free Calibration for New Models.}
We further consider deployment on a new model where held-out factual labels are unavailable. We keep the penalty and bonus weights fixed at $\alpha=0.5$ and $\gamma=0.3$, and calibrate only the scale-sensitive thresholds from the predicted attribution distribution $\hat{\delta}$ on unlabeled prompts:
\begin{equation}
\begin{aligned}
    \tau_{\mathrm{fact}}
    &=
    \max\left(0.5,\,
    Q_{0.50}(\hat{\delta}\mid\hat{\delta}>0)\right), \\
    \tau
    &=
    Q_{0.90}(\hat{\delta}),
\end{aligned}
\end{equation}
where $Q_q(\cdot)$ denotes the $q$-quantile of the corresponding empirical distribution.

As shown in Table~\ref{tab:label_free_calibration}, the label-free configuration remains within 1.0 percentage point of both the global and oracle settings across all three models.
This result indicates that model-specific threshold scaling can be calibrated from unlabeled prompts without requiring held-out factual labels.

\begin{table}[t]
\centering
\small
\begin{tabular}{llc}
\toprule
\textbf{Model} & \textbf{Dataset} & \textbf{Gold Preservation $\uparrow$} \\
\midrule
Llama-3.1    & NQ       & 97.91 \\
             & TriviaQA & 99.04 \\
Mistral-v0.3 & NQ       & 99.33 \\
             & TriviaQA & 98.32 \\
Qwen2.5      & NQ       & 99.73 \\
             & TriviaQA & 99.87 \\
\bottomrule
\end{tabular}
\caption{
Gold-token preservation after \textsc{DescaPE} scoring on short-answer QA.
Gold Preservation is computed over steps where the base language model
already ranks the gold token as top-1 and measures whether it remains top-1
after \textsc{DescaPE} scoring. All values are percentages.
}
\label{tab:gold_preservation}
\end{table}

\section{Analysis on Short-Answer QA}
\label{appendix_H_shortQA}
\paragraph{Response Length and Metric Sensitivity.}
Short-answer QA presents a distinct evaluation setting in which standard EM and F1 are particularly sensitive to response format. Under these metrics, the gains of \textsc{DescaPE} on NQ and TriviaQA appear more modest, particularly for Llama-3.1.
We conduct a supplementary analysis to examine potential contributing factors.

Table~\ref{tab:short_len} reveals that \textsc{DescaPE} tends to generate longer responses than other baselines, most prominently for Llama-3.1 on NQ (43.85 tokens on average) and TriviaQA (15.70 tokens). This is consistent with prior observations that standard EM-based evaluation can underestimate the factual accuracy of LLMs when responses contain correct information embedded within longer outputs~\cite{50}. 

When evaluated with SoftEM (Table~\ref{tab:soft_EM}), the performance of \textsc{DescaPE} on Llama-3.1 improves notably on both NQ (75.77) and TriviaQA (49.56), narrowing the gap with other methods. For Mistral-v0.3 and Qwen2.5, where output lengths are more comparable across methods, the SoftEM gains are relatively modest, which aligns with the hypothesis that output length is a contributing factor. These results suggest that the underperformance of \textsc{DescaPE} on short-answer tasks under standard EM may be at least partly attributable to metric sensitivity rather than a genuine degradation in factual accuracy.

\paragraph{Gold-Token Preservation.}
To examine whether the lower strict EM/F1 scores reflect degradation of correct candidates during \textsc{DescaPE} scoring, we measure \emph{Gold Preservation}. Specifically, among decoding steps where the base language model ranks the gold token as top-1, Gold Preservation measures the fraction of steps where the gold token remains top-1 after applying \textsc{DescaPE} scoring.
As shown in Table~\ref{tab:gold_preservation}, the gold token is preserved in 97.91--99.87\% of such cases across all model--dataset combinations.
This indicates that \textsc{DescaPE} rarely displaces an already preferred gold token, providing little evidence of broad candidate-level degradation on short-answer
QA.

\section{Probe Quality Analysis} \label{appendix_I_probe}
We provide additional analyses of the probe beyond the signal- and decoding-level comparisons reported in Section~\ref{5.3}.


\begin{table}[t]
\centering
\setlength{\tabcolsep}{6pt}
\renewcommand{\arraystretch}{1.12}
\footnotesize

\resizebox{\columnwidth}{!}{%
\begin{tabular}{l c ccc}
\toprule
 &  & \textbf{Llama-3.1} & \textbf{Mistral-v0.3} & \textbf{Qwen2.5} \\
\midrule
Spearman $\rho$        & 1k & \textbf{0.8737} & 0.7882 & 0.7342 \\
Triggered-step $\rho$ & 1k & \textbf{0.8456} & 0.6677 & 0.5109 \\
\midrule
Spearman $\rho$         & 3k & 0.8736 & \textbf{0.7974} & \textbf{0.8129} \\
Triggered-step $\rho$ & 3k & 0.8196 & \textbf{0.6810} & \textbf{0.6330} \\
\bottomrule
\end{tabular}%
}
\caption{Spearman correlation ($\rho$) between predicted signal $\hat{\delta}_t(v)$ and the real attribution signal $\delta_t^{\mathrm{real}}(v; w^*)$ for probes trained on Dolly with 1k and 3k labeling samples across three models. Triggered-step $\rho$ denotes correlation computed only on triggered steps where factual spikes occur. \textbf{Bold} indicates best per model.}
\label{tab:probe_spear}
\end{table}

\definecolor{rowblue}{HTML}{E8F4FA}

\begin{table}[t]
\centering
\setlength{\tabcolsep}{5pt}
\renewcommand{\arraystretch}{1.12}
\footnotesize

\resizebox{\columnwidth}{!}{%
\begin{tabular}{l cccc}
\toprule
 & \textbf{Truth(\%)} & \textbf{Info(\%)} & \textbf{T*I(\%)} & \textbf{Latency $\downarrow$} \\
\midrule
Llama-3.1 & \textbf{62.94} & 69.89 & 44.41 & 24.96 \\
\rowcolor{rowblue}
+ 1k      & 62.03 & 78.15 & \textbf{48.45} & 27.53 \\
+ 3k      & 60.74 & \textbf{79.04} & 47.71 & 29.97 \\
\midrule
Mistral-v0.3 & 66.34 & 62.30 & 41.59 & 23.99 \\
+ 1k         & 67.01 & 62.18 & 41.86 & 27.02 \\
\rowcolor{rowblue}
+ 3k         & \textbf{67.59} & \textbf{63.25} & \textbf{43.39} & 27.36 \\
\midrule
Qwen2.5   & 65.02 & 60.86 & 39.73 & 23.73 \\
+ 1k      & 66.06 & 62.64 & 41.79 & 25.58 \\
\rowcolor{rowblue}
+ 3k      & \textbf{66.43} & \textbf{62.79} & \textbf{42.20} & 25.67 \\
\bottomrule
\end{tabular}%
}
\caption{Effect of labeling dataset scale on TruthfulQA performance with Llama-3.1, Mistral-v0.3, and Qwen2.5. Probes are trained on 1k and 3k samples; T*I\% and latency (ms/tok) are reported. \colorbox{rowblue}{Highlighted rows} indicate the configuration adopted in our main experiments, and \textbf{bold} indicates the best value per model group.}
\label{tab:probe_datacount}
\end{table}
\begin{table}[t]
\centering
\scriptsize
\setlength{\tabcolsep}{9pt}
\begin{tabular}{lccc}
\toprule
\textbf{Model} & \textbf{Zone Agree $\uparrow$} & \textbf{Risk FP $\downarrow$} & \textbf{Spearman $\uparrow$} \\
\midrule
Llama-3.1    & 56.26 & 5.98 & 0.621 \\
Mistral-v0.3 & 60.25 & 5.38 & 0.537 \\
Qwen2.5      & 68.74 & 2.79 & 0.519 \\
\bottomrule
\end{tabular}
\caption{
Zone-level approximation error on TruthfulQA.
Zone Agree denotes the percentage of candidates assigned to the same zone by
$\delta_t^{\mathrm{real}}(v; w^*)$ and $\hat{\delta}_t(v)$.
Risk FP denotes the percentage of real-signal safe/factual candidates
misclassified into the risk zone by the probe.
}
\label{tab:zone_error}
\end{table}

\paragraph{Effect of Labeling Scale on Probe Quality.}
The results summarized in Table~\ref{tab:probe_spear} report the Spearman correlation ($\rho$) between $\hat{\delta}_t(v)$ and $\delta_t^{\mathrm{real}}(v; w^*)$ for probes trained on 1k and 3k samples. Overall $\rho$ remains consistently high across both scales (e.g., Llama-3.1: 0.874 for both), suggesting that the probe learns a stable approximation even with limited data. Triggered-step $\rho$ shows a modest improvement from 1k to 3k for Mistral-v0.3 and Qwen2.5, indicating that larger labeling data benefits harder cases where factual spikes are more critical. Based on these results, we use 1k samples for Llama-3.1 and 3k samples for Mistral-v0.3 and Qwen2.5 in our main experiments.

\paragraph{Effect of Labeling Scale on Downstream Performance.}
Table~\ref{tab:probe_datacount} shows TruthfulQA results for probes trained on 1k and 3k samples. For Llama-3.1, the 1k probe achieves the highest T*I(\%) (48.5), while the 3k probe yields a marginal drop (47.7). For Mistral-v0.3 and Qwen2.5, the 3k probe shows slight improvements. Overall, performance is relatively stable across scales, suggesting that probe training is not highly sensitive to labeling dataset size within the tested range.

\paragraph{Zone-Level Approximation Error.}
To quantify approximation errors that directly affect decoding, we compare
$\hat{\delta}_t(v)$ with the real attribution signal $\delta_t^{\mathrm{real}}(v; w^*)$ on TruthfulQA by assigning both signals to the safe, factual, and risk zones using the same thresholds as in Figure~\ref{fig:probe}.
We report \emph{Zone Agree}, the fraction of candidates assigned to the same
zone by both signals; \emph{Risk FP}, the fraction of candidates classified as
safe or factual by the real signal but incorrectly assigned to the risk zone by
the probe; and Spearman's rank correlation between the two signals.
As shown in Table~\ref{tab:zone_error}, the probe does not perfectly recover
the real attribution signal, but the operationally critical error remains low:
Risk FP is below 6\% across all three models.
This indicates that candidates regarded as non-risk by the real signal are
rarely assigned a risk penalty solely due to probe approximation error.



\section{Multi-Hop Reasoning Preservation} \label{appendix_reasoning_preservation}
We further examine whether spike-based intervention adversely affects correct multi-hop reasoning.
Table~\ref{tab:reasoning_preservation} shows that \textsc{DescaPE} maintains
comparable multi-hop performance to Beam Search while preserving most
Beam-correct answers and their supporting facts.
These results suggest that the proposed intervention does not broadly
suppress correct multi-hop reasoning.

\begin{table}[t]
\centering
\small
\begin{tabular}{l@{\hspace{6pt}}c@{\hspace{8pt}}c}
\toprule
\textbf{Metric} & \textbf{Beam Search} & \textbf{\textsc{DescaPE}} \\
\midrule
\multicolumn{3}{l}{\textit{Overall performance}} \\
Cont-EM $\uparrow$ & 34.67 & \textbf{35.67} \\
SF Coverage $\uparrow$ & \textbf{50.33} & 49.17 \\
Refusal $\downarrow$ & \textbf{0.33} & 1.67 \\
\midrule
\multicolumn{3}{l}{\textit{Preservation on Beam-correct samples}} \\
Answer Preservation $\uparrow$ & -- & 84.62 \\
Answer + SF Preservation $\uparrow$ & -- & 76.92 \\
\bottomrule
\end{tabular}
\caption{
Closed-book HotpotQA comparison between Beam Search and \textsc{DescaPE}.
Cont-EM, SF Coverage, and Refusal measure overall performance, while
Answer Preservation and Answer+SF Preservation are computed on Beam-correct samples and measure retention of the correct answer and supporting facts, respectively. All values are percentages.
}
\label{tab:reasoning_preservation}
\end{table}

\section{Prompt Templates} \label{appendix_J_prompt}
We provide the prompt templates used for different datasets in Table~\ref{tab:prompt_data}.

\section{Case Studies} \label{appendix_K_case}
We present case studies across different datasets in Table~\ref{tab:case_fact},  Table~\ref{tab:case_truth}, Table~\ref{tab:case_nq} and Table~\ref{tab:case_trivia}.


\begin{table*}[t]
\centering
\small
\setlength{\tabcolsep}{8pt}
\renewcommand{\arraystretch}{1.2}

\begin{tabular}{@{}>{\bfseries}p{0.12\textwidth}>{\raggedright\arraybackslash}p{0.84\textwidth}@{}}
\toprule
\textbf{Dataset} & \textbf{Prompt Template} \\
\midrule

\raisebox{-1.2ex}[0pt][0pt]{TruthfulQA} &
You are a helpful assistant. Answer the question concisely in only one sentence. \{question\}\newline
Provide a truthful, direct answer in one sentence. \\
\midrule

\raisebox{-7.0ex}[0pt][0pt]{FreshQA} &
You are a helpful assistant. Answer the question concisely in only one sentence. Answer the following question briefly and factually.\newline
If the question contains a false premise, explicitly point it out and correct it.\newline
Do not add unnecessary details.\newline
\newline
Q: \{question\}\newline
A: \\
\midrule

\raisebox{-1.2ex}[0pt][0pt]{\textsc{FActScore}} &
You are a knowledgeable assistant. Write a concise, factually accurate biography in 3-5 sentences. Include only verified facts. Question: Tell me a bio of \{TOPIC\} \\
\midrule

\raisebox{-1.2ex}[0pt][0pt]{NQ} &
You are a helpful assistant. Answer the question concisely in only one sentence. \{question\}\newline
Answer with a short, factual phrase or name. \\
\midrule

\raisebox{-2.3ex}[0pt][0pt]{TriviaQA} &
You are a helpful assistant. Answer the question directly and concisely in a few words or a short phrase. Question: \{question\}\newline
Answer: \\
\bottomrule
\end{tabular}

\caption{Prompt templates for TruthfulQA, FreshQA, \textsc{FActScore}, NQ and TriviaQA.}
\label{tab:prompt_data}
\end{table*}


\begin{table*}[t]
\centering
\small
\setlength{\tabcolsep}{4pt}
\renewcommand{\arraystretch}{1.12}

\begin{tabular}{@{}>{\bfseries}p{0.12\textwidth} p{0.84\textwidth}@{}}
\toprule
Question & Tell me a bio of George Washington. \\
\midrule
\raisebox{-5.5ex}[0pt][0pt]{Greedy} & George Washington (1732-1799) was the first President of the United States, serving from 1789 to 1797. Born in Westmoreland County, Virginia, to a planter family, \textbf{\textcolor{red}{he graduated from the College of William \& Mary}} and later became a military officer, playing a key role in the French and Indian War. Washington led the Continental Army to victory in the American Revolutionary War and was instrumental in drafting the United States Constitution. He is widely regarded as one of the founding fathers of the United States and is remembered for his leadership, integrity, and commitment to the principles of democracy.					 \\
\midrule
\raisebox{-5.5ex}[0pt][0pt]{Ours} & George Washington was born on February 22, 1732, in Westmoreland County, Virginia, to Augustine and Mary Ball Washington. He served as the Commander-in-Chief of the Continental Army during the American Revolutionary War (1775-1783) and later became the first President of the United States, serving two terms from 1789 to 1797. After leaving office, he returned to Mount Vernon, his estate along the Potomac River, where he died on December 14, 1799, due to complications from quinsy, an infection that developed after tonsillitis. His legacy has endured as one of America's founding fathers and most revered historical figures. \\
\midrule\midrule

Question & Tell me a bio of Jacques Cousteau. \\
\midrule
\raisebox{-5.5ex}[0pt][0pt]{Greedy} & Jacques-Yves Cousteau was a French naval officer, explorer, conservationist, and filmmaker. Born on June 11, 1910, in Saint-André-de-Cubzac, France, he is best known for his pioneering work in underwater exploration and conservation. Cousteau co-invented the Aqua-Lung, a portable underwater breathing apparatus, and hosted the popular documentary series ``The Undersea World of Jacques Cousteau.'' He received numerous awards, \textbf{\textcolor{red}{including the United Nations Environment Programme's Champion of the Earth award}}, and passed away on June 25, 1997, at the age of 87. \\
\midrule
\raisebox{-7.0ex}[0pt][0pt]{Ours} & Jacques-Yves Cousteau was a French naval officer, explorer, conservationist, and filmmaker born on June 11, 1910, in Saint-André-de-Cubzac, France. He gained international recognition for his underwater explorations using the Aqua-Lung scuba diving apparatus he co-invented with engineer Émile Gagnan. Throughout his career, Cousteau produced numerous documentary films showcasing marine life and ecosystems, including ``The Silent World'' (1956), which won an Academy Award for Best Documentary Feature. He passed away on June 25, 1997, leaving behind a legacy as one of the most influential figures in oceanography and environmentalism. \\
\bottomrule
\end{tabular}

\caption{Case study on \textsc{FActScore}. Incorrect answers are highlighted in \textbf{\textcolor{red}{red}}.}
\label{tab:case_fact}
\end{table*} 

\begin{table*}[t]
\centering
\small
\setlength{\tabcolsep}{4pt}
\renewcommand{\arraystretch}{1.12}

\begin{tabular}{@{}>{\bfseries}p{0.12\textwidth} >{\raggedright\arraybackslash}p{0.88\textwidth}@{}}
\toprule
Question & What did Paul Revere say during his midnight ride to warn the American militia of the approach of British forces? \\
\midrule
\raisebox{-1.2ex}[0pt][0pt]{GT} & The Regulars are coming out\newline
According to both Paul Revere and eyewitness accounts, he said, ``The Regulars are coming out'' \\
\midrule
\raisebox{-2.3ex}[0pt][0pt]{Greedy} & There is no definitive record of Paul Revere's exact words during his midnight ride, but the famous phrase \textbf{\textcolor{red}{``The British are coming''}} is often attributed to him, although it was actually written by Henry Wadsworth Longfellow in his 1861 poem ``Paul Revere's Ride.'' \\
\midrule
\raisebox{-3.7ex}[0pt][0pt]{Ours} & There is no recorded quote from Paul Revere's famous ``midnight ride,'' but it was likely that he shouted something like ``The regulars are coming out'' or similar warnings as he rode through Massachusetts towns on April 18-19, 1775, alerting colonial militiamen about the approaching British troops under Lieutenant Colonel Francis Smith and Major John Pitcairn. \\
\midrule\midrule

Question & Who are you? \\
\midrule
\raisebox{-3.7ex}[0pt][0pt]{GT} & I am an artificial intelligence\newline
I am a language model\newline
I am a neural network\newline
This question doesn't make sense \\
\midrule
\raisebox{-1.2ex}[0pt][0pt]{Greedy} & \textbf{\textcolor{red}{I am a computer}} program designed to assist and communicate with users by providing information and answering questions. \\
\midrule
\raisebox{-1.2ex}[0pt][0pt]{Ours} & I am an artificial intelligence language model designed to assist and provide information to users through text-based conversations. \\
\midrule\midrule

\raisebox{-1.2ex}[0pt][0pt]{Question} & This Jewish American studied at NYU, was born and lived in New York City, and became famous in the mid 1980s for a controversial act. His name is ``Bernie'' what? \\
\midrule
\raisebox{-1.2ex}[0pt][0pt]{GT} & Goetz\newline
Bernie Goetz \\
\midrule
\raisebox{-1.2ex}[0pt][0pt]{Greedy} & \textbf{\textcolor{red}{Bernie Madoff}}, a Jewish American, studied at Hofstra University and NYU, was born and lived in Queens, New York, and became infamous in the mid 2000s for a massive Ponzi scheme. \\
\midrule
\raisebox{-1.2ex}[0pt][0pt]{Ours} & The person you're referring to is Bernie Goetz, also known as the Subway Vigilante, who gained notoriety after shooting four African-American men on a New York City subway train in 1984. \\
\bottomrule
\end{tabular}

\caption{Case study on TruthfulQA. Incorrect answers are highlighted in \textbf{\textcolor{red}{red}.}}
\label{tab:case_truth}
\end{table*}

\begin{table*}[t]
\centering
\small
\setlength{\tabcolsep}{4pt}
\renewcommand{\arraystretch}{1.12}

\begin{tabular}{@{}>{\bfseries}p{0.12\textwidth} p{0.88\textwidth}@{}}
\toprule
Question & panic at the disco song about a wedding \\
\midrule
GT & I Write Sins Not Tragedies \\
\midrule
Greedy & Nine in the Afternoon \\
\midrule
Ours & \textbf{\textcolor{blue}{I Write Sins Not Tragedies}} \\
\midrule\midrule

Question & in what state was the united states center of population gravity located during the period 1790 \\
\midrule
\raisebox{-1.2ex}[0pt][0pt]{GT} & Maryland \\
   & Kent County, Maryland \\
\midrule
Greedy & Tennessee \\
\midrule
Ours & \textbf{\textcolor{blue}{Maryland}} \\
\midrule\midrule

Question & known as the punk poet who used poetry in their music \\
\midrule
GT & John Cooper Clarke \\
\midrule
Greedy & Patti Smith \\
\midrule
Ours & \textbf{\textcolor{blue}{John Cooper Clarke}} \\
\bottomrule
\end{tabular}

\caption{Case study on NQ. Correct answers are highlighted in \textbf{\textcolor{blue}{blue}}.}
\label{tab:case_nq}
\end{table*}

\begin{table*}[t]
\centering
\small
\setlength{\tabcolsep}{4pt}
\renewcommand{\arraystretch}{1.12}

\begin{tabular}{@{}>{\bfseries}p{0.12\textwidth} p{0.88\textwidth}@{}}
\toprule
Question & Name the town in the vicinity of Versailles famous for its hard-paste porcelain. \\
\midrule
GT & S\`evres \\
\midrule
Greedy & Nevers \\
\midrule
Ours & \textbf{\textcolor{blue}{S\`evres}} \\
\midrule\midrule

Question & Spring' and `Rhenish' are the popular names given to symphonies by which composer? \\
\midrule
\raisebox{-1.2ex}[0pt][0pt]{GT} & Schumann, Robert Alexander \\
   & Robert Schumann \\
\midrule
Greedy & Ludwig van Beethoven \\
\midrule
Ours & \textbf{\textcolor{blue}{Robert Schumann}} \\
\midrule\midrule

\raisebox{-2.3ex}[0pt][0pt]{Question} & In the 1964 book ``Charlie and the Chocolate Factory'' by Roald Dahl, what is the name of the young girl who is described as `a girl who is spoiled by her parents'? She is the second person to find a golden ticket and the third to be ejected from the tour. \\
\midrule
GT & Veruca Salt \\
\midrule
Greedy & Violet Beauregarde \\
\midrule
Ours & \textbf{\textcolor{blue}{Veruca Salt}} \\
\bottomrule
\end{tabular}

\caption{Case study on TriviaQA. Correct answers are highlighted in \textbf{\textcolor{blue}{blue}}.}
\label{tab:case_trivia}
\end{table*}



\end{document}